\documentclass{article} 
\usepackage{iclr2026_conference,times}

\usepackage{amsmath,amsfonts,bm}

\def\eqref#1{equation~\ref{#1}}

\def\1{\bm{1}}

\DeclareMathAlphabet{\mathsfit}{\encodingdefault}{\sfdefault}{m}{sl}
\SetMathAlphabet{\mathsfit}{bold}{\encodingdefault}{\sfdefault}{bx}{n}

\usepackage{hyperref}
\usepackage{url}
\usepackage{graphicx}
\usepackage{wrapfig}

\title{Functional Embeddings Enable Aggregation of Multi-Area SEEG Recordings Over\\ Subjects and Sessions}

\author{Sina Javadzadeh\\
Department of Biomedical Engineering, University of California Irvine\\
California, USA \\
\texttt{javdzas@uci.edu} \\
\AND
R. Soroushmojdehi, S. Alireza~Seyyed Mousavi,~\&M. Asadi\\
Department of Electrical Engineering, University of California Irvine \\
California, USA \\
\texttt{\{rsoroush,sseyyedm,mehrnaza\}@uci.edu} \\
\AND
S. Abe \\
Department of Neurology, Rady Children's Health \\
California, USA \\
\texttt{sumiko.abe@choc.org}\\
\AND
Terence D. Sanger  \\
Department of Electrical Engineering, University of California Irvine \\
Department of Neurology, Rady Children's Health \\
California, USA \\
\texttt{tsanger@uci.edu} \\
}

\iclrfinalcopy 
\begin{document}

\maketitle

\pagestyle{fancy}
\fancyhf{} 
\lhead{\textit{Preprint. Submitted to ICLR 2026.}}
\rhead{} 
\cfoot{\thepage} 

\begin{abstract}
Aggregating intracranial recordings across subjects is challenging since electrode count, placement, and covered regions vary widely. Spatial normalization methods like MNI coordinates offer a shared anatomical reference, but often fail to capture true functional similarity, particularly when localization is imprecise; even at matched anatomical coordinates, the targeted brain region and underlying neural dynamics can differ substantially between individuals. We propose a scalable representation-learning framework that (i) learns a subject-agnostic functional identity for each electrode from multi-region local field potentials using a Siamese encoder with contrastive objectives, inducing an embedding geometry that is locality-sensitive to region-specific neural signatures, and (ii) tokenizes these embeddings for a transformer that models inter-regional relationships with a variable number of channels. We evaluate this framework on a 20-subject dataset spanning basal ganglia–thalamic regions collected during flexible rest/movement recording sessions with heterogeneous electrode layouts. The learned functional space supports accurate within-subject discrimination and forms clear, region-consistent clusters; it transfers zero-shot to unseen channels. The transformer, operating on functional tokens without subject-specific heads or supervision, captures cross-region dependencies and enables reconstruction of masked channels, providing a subject-agnostic backbone for downstream decoding. Together, these results indicate a path toward large-scale, cross-subject aggregation and pretraining for intracranial neural data where strict task structure and uniform sensor placement are unavailable. 

\end{abstract}

\section{Introduction}
Building models that generalize across subjects in neuroscience requires representations that remain stable despite variability in data acquisition. Intracranial neural recordings lack this stability: electrode locations, counts, sampling, and coverage differ across individuals, reflecting clinical needs rather than standardized layouts. Without a shared representational system, cross-subject aggregation is unreliable, limiting scalable modeling and clinical translation.

These difficulties are magnified in emerging invasive multi-region datasets, where electrodes simultaneously capture activity from distinct deep-brain structures\citep{sanger2018case,kasiri2024local,hernandez2023effect,hernandez2023globus,kasiri2023correlated,cimbalnik2025human,parmigiani2022simultaneous,shahabi2023multilayer,wu2023deep,wu2024channel,kasiri2025optimized}. Such recordings are uniquely valuable for studying inter-regional communication, yet their heterogeneity makes them especially challenging to align. In practice, two obstacles dominate:

\textbf{Anatomical variability and inconsistent electrode coverage.} Electrode placement can vary significantly across subjects \citep{starr2006microelectrode,schonecker2015postoperative}. Standard atlas-based alignment (e.g., MNI \citep{talairach1988co}) assumes spatial correspondence implies functional similarity, yet recordings at matched coordinates often capture divergent functional roles\citep{nestor2014coordinate}, and in some cases completely different brain regions \citep{ashkan2007variability, daniluk2010assessment}.

\textbf{Multi-area recordings.} Modern DBS procedures now routinely sample from several basal ganglia and thalamic nuclei simultaneously \citep{liker2024stereotactic,sanger2018pediatric}. These datasets offer unprecedented opportunities to study inter-regional communication, but their heterogeneity magnifies the alignment problem: each region expresses distinct yet interdependent functional signatures.

Attempts to scale datasets, inspired by trends in language modeling \citep{wiggins2022opportunities,kaplan2020scaling}, are in many cases limited to multi-session data within a single subject and region \citep{pandarinath2018inferring,farshchian2018adversarial,karpowicz2025stabilizing,degenhart2020stabilization}, or combined, but through post-hoc latent alignment \citep{safaie2023preserved,ma2023using}, which precludes unified end-to-end training across multi-session, multi-subject datasets. 

Transformer-based models have begun to address this by training foundation models for EEG decoding \citep{wan2023eegformer,ding2025emt,cui2024neuro}, but they typically assume fixed, high-density electrode grids and large labeled datasets. Extending these ideas to invasive recordings is harder: spatial coverage is sparse, contact locations vary across patients, and anatomical localization can be imprecise. Consequently, many approaches rely on atlas-based alignment \citep{mentzelopoulos2024seegnificant,chau2025popT}, which inherits the limitations of standardized coordinate systems noted above. This motivates our central hypothesis:
\textbf{Neural signals can be aligned across subjects more reliably by their functional characteristics than by their anatomical coordinates.}

We operationalize this hypothesis in \emph{FunctionalMap}, a framework that treats “function” as the coordinate system. A Siamese encoder \citep{hadsell2006dimensionality,bromley1993signature},  trained with contrastive objectives learns subject-agnostic embeddings of short neural segments, mapping signals from the same brain region close together in latent space regardless of electrode placement. These embeddings serve as functional identifiers in tokens for a transformer \citep{vaswani2017attention, dosovitskiy2020image}, which models inter-regional dependencies across variable channel sets without subject IDs or subject-specific heads. 
To our knowledge, we are the first to learn a subject-invariant, region-structured functional coordinate system from invasive, multi-region human LFPs and to use it for end-to-end training on a unified multi-session, multi-subject dataset.
\textbf{Our contributions:}
\textbf{Functional coordinates.} We introduce data-driven embeddings that capture region-specific dynamics, yielding a subject-agnostic functional identity that replaces unreliable atlas-based alignment.
\textbf{Unified cross-subject modeling.} Embedding-informed tokenization lets a single transformer aggregate heterogeneous intracranial datasets—irregular electrode layouts, multiple regions, without per-subject models.
\textbf{Empirical validation.} On a 20-subject multi-region intracranial LFP dataset functional embedding was able to cluster regions and transferred zero-shot to held-out channels. when ablating coordinate systems on a 11-subject dataset, functional embeddings \emph{significantly improve} reconstruction compared to MNI-coordinate alignment.
Together, these results suggest that data-derived coordinate systems provide the missing backbone for large-scale modeling of invasive human recordings, opening the door to subject-agnostic pretraining, cross-subject decoding, and scalable representation learning in neuroscience. 
To make this work reproducible, all codes used in this study are available at \url{https://github.com/ICLR-Functional-Embedding/ICLR2026_Functional_Map}. 
\vspace{-2mm}
\section{Related Work} 
\paragraph{Transformers for neural population modeling}
Transformer architectures have advanced modeling of neural population activity across spikes and field potentials, including NDT \citep{ye2021NDT1,ye2023NDT2}, STNDT \citep{le2022stndt}, and large-scale decoding frameworks \citep{azabou2023unified, ding2025emt,cui2024neuro}. These methods excel at spatiotemporal sequence modeling but generally assume stable channel identities/dense coverage or use neuronal level spiking activity, and do not provide mechanisms for aligning heterogeneous electrode sets across subjects. Our method complements these works: we retain transformer scalability while injecting \emph{functional embeddings} that align signals across subjects and regions prior to aggregation.
\vspace{-6mm}
\paragraph{Neural decoding with SEEG}
Recent SEEG studies demonstrate decoding for speech and other behaviors from intracranial recordings \citep{angrick2021real,meng2021identification,petrosyan2022speech,wu2024channel,asadi2025bace}. These systems are often optimized per dataset or subject and rely on task-specific supervision. In contrast, we aim to first learn subject-/region-agnostic \emph{functional} coordinates without behavioral labels, then plug these coordinates into transformers to scale across subjects and tasks.
\paragraph{Cross-subject SEEG decoding under electrode variability}
SEEG cohorts exhibit heterogeneous electrode numbers and placements across subjects, complicating multi-subject integration. \citet{wang2023brainbert} proposes a subject and electrode agnostic neural representations using transformers trained on masked spectrogram reconstruction, our goal is not generic contextualization but functional alignment: we learn region-structured embeddings via contrastive objective, addressing the well-known gap between anatomical coordinates and functional identity.
\citet{mentzelopoulos2024seegnificant} address this by tokenizing per-electrode activity, injecting 3D coordinates, and performing attention across time and electrodes with subject-specific heads. notably, \citet{mentzelopoulos2024seegnificant} found no significant performance loss when ablating positional encoding using MNI coordinates. \citet{chau2025popT} pretrain a Population Transformer that aggregates arbitrary, spatially sparse channel ensembles via self-supervision, improving downstream decoding and few-shot transfer. In contrast, our approach learns \emph{functional} embeddings directly from neural dynamics and uses them as universal coordinates for cross-subject aggregation, avoiding reliance on atlas coordinates.

\vspace{-2mm}
\section{Method}

\begin{figure}[!b]\center{\includegraphics[width=1\textwidth,trim={0cm 5cm 0cm 5cm},clip=true] {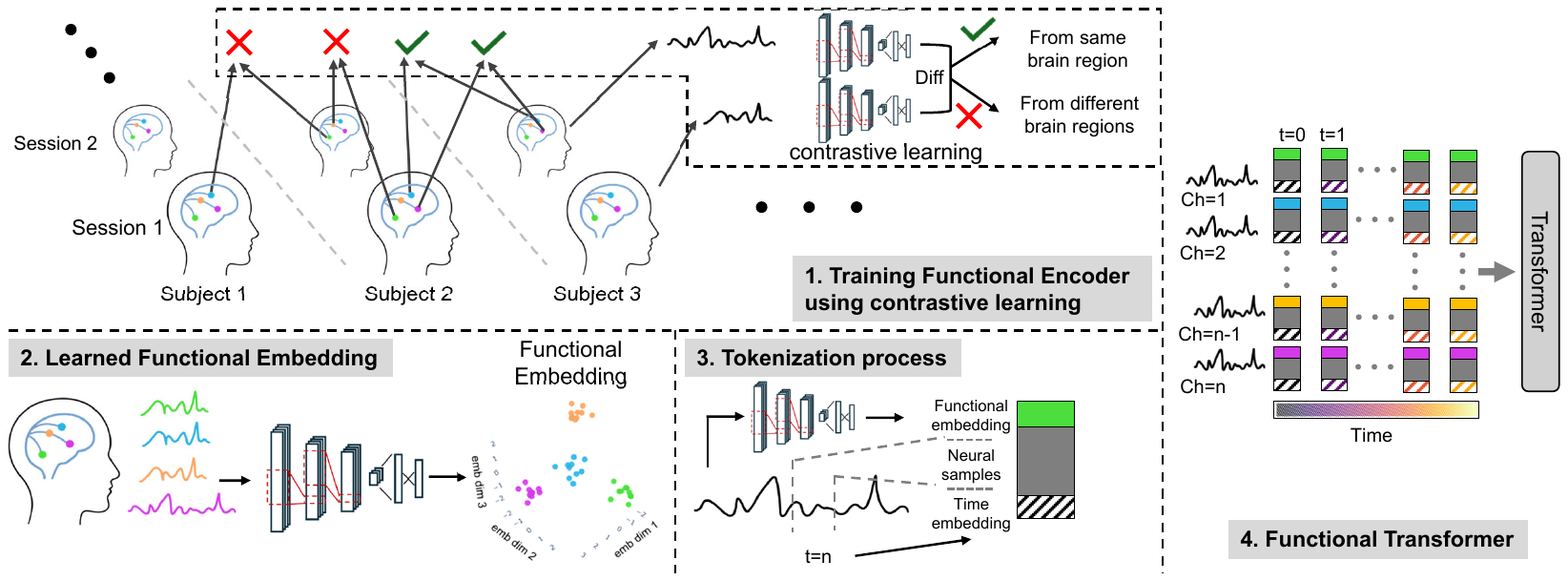}}
	\caption{
		\textit{Overview of Approach.} 1: Contrastive learning framework used to train the CNN-based functional encoder via a Siamese architecture, encouraging signals from the same brain region to map nearby in latent space. 2: Learned functional embeddings produced by the trained encoder, showing clustering by brain region. 3: Incorporating functional embeddings into the tokenization scheme. 4: Functional Transformer architecture that enables scalable modeling across subjects and sessions .
		}
	\label{fig1}
\end{figure}

We propose a scalable framework for learning functionally meaningful representations from multi-region human intracranial neural recordings, designed to generalize across subjects and sessions with variable electrode configurations (Figure~\ref{fig1}). Our approach first constructs a functional embedding space using a contrastive learning trained to map neural signals from the same brain region closer together in latent space. These learned embeddings are then used to tokenize input data and condition a transformer-based architecture for downstream tasks such as time-series reconstruction. This two-stage framework supports subject-agnostic modeling by capturing data-driven regional identity based on neural dynamics rather than anatomical location (i.e., MNI coordinates).

\paragraph{Data Collection and Preprocessing}
Our dataset comprises intracranial local field potential (LFP) recordings from 20 subjects with dystonia who underwent deep brain stimulation (DBS) procedure. Stereo-EEG (SEEG) electrodes were implanted in multiple basal ganglia and thalamic nuclei, including the internal segment of the globus pallidus (GPi), the subthalamic nucleus (STN), the ventralis oralis thalamic nucleus (VO), Ventral Anterior thalamic nucleus (VA), ventral intermediate thalamic nucleus (VIM), pedunculopontine nucleus (PPN), and substantia nigra pars reticulata (SNr), with electrode targets selected based on clinical criteria. In total, the dataset includes over 442.86 electrode-hours of neural recordings. For transformer related analysis only 11 out of 20 subjects were used based on the availability of their MNI data.  Further details on datasets, electrode locations, and preprocessing steps are available in Appendices~\ref{app:electrode} and~\ref{A_preprocessing}.

\subsection{Functional Embedding Network}
To address anatomical variability and heterogeneous electrode coverage across subjects, we learn a 32-dimensional (32-D) functional embedding for each recording channel using a Siamese neural network trained with contrastive loss. The objective is to construct a latent space in which signals from the same brain region, regardless of subject or session, are mapped close together (locality-sensitive functional map), thereby enabling data aggregation across sessions and individuals.

The network receives pairs/groups of LFP segments as input and minimizes a contrastive loss function that brings embeddings of positive pairs (same region) closer while pushing negative pairs (different regions) apart. The encoder architecture is a lightweight convolutional neural network (CNN) with batch normalization, GELU activation function \citep{hendrycks2016gaussian}, and dropout \citep{srivastava2014dropout}. We implemented two contrastive methods: 

\textbf{Method 1: Pairwise Siamese Contrastive (PSC)}\\
Let $f_{\theta}:\mathbb{R}^{T}\!\to\!\mathbb{R}^d$ be the encoder that maps a 10~s segment $x$ to an embedding $z=f_\theta(x)$.
Given a balanced set of pairs $\{(x_i,x_j,y_{ij})\}$ with $y_{ij}\!\in\!\{0,1\}$ ($0$ similar: same region; $1$ dissimilar: different regions), let $z_i$ and $z_j$ be their embeddings and $d_{ij}=\|z_i-z_j\|_2$. Here, $\mathcal{B}$ denotes the set of index pairs in the current mini-batch. We minimize the standard contrastive loss with margin $m>0$:
\[
\mathcal{L}_{\text{pair}}
\;=\;
\frac{1}{|\mathcal{B}|}\sum_{(i,j)\in\mathcal{B}}
\Big[(1-y_{ij})\,d_{ij}^{2}
\;+\;
y_{ij}\,\big(\max(0,\,m-d_{ij})\big)^{2}\Big].
\]
 In the case of similar pairs ($y_{ij}=0$) only the first term in the summation is active making the loss proportional to the distance of embeddings ($d_{ij}^{2}$), penalizing long distances in the embedding space from same-source sample pairs. In the case of dissimilar pairs ($y_{ij}=1$), the second term in the summation is active making the loss function proportional to the closeness of embeddings \big($(m-d_{ij})^{2}$\big), if distance is bigger than margin ($m=0.5$), the loss becomes zero. This encourages dissimilar pair embeddings to be at least $m$ apart\citep{hadsell2006dimensionality}.

\textbf{Method 2: Modified SupCon (MSC)}\\
We utilized a modified Supervised Contrastive approach (SupCon; \citet{khosla2020supervised}) in a single-view, projection-free setting. This method uses multi-positive InfoNCE \citep{oord2018representation} with label-defined positives. Let $f_\theta:\mathbb{R}^{T}\!\to\!\mathbb{R}^d$ be the encoder (same as PSC) producing an embedding $z_i=f_\theta(x_i)$ for segment $x_i$ with region label $y_i$.
For a mini-batch $\mathcal{B}=\{(x_i,y_i)\}_{i=1}^{N}$, define the positive set for anchor $i$ as
$\mathcal{P}(i)=\{\,p\in\{1,\dots,N\}\setminus\{i\}:~y_p=y_i\,\}$.
Using inner-product similarity $s_{ij}=z_i^\top z_j$ and temperature $\tau>0$ in our case $\tau=0.2$:
\[
\mathcal{L}_{\text{sup}}
=\frac{1}{N}\sum_{i=1}^{N}
\left(
-\frac{1}{|\mathcal{P}(i)|}
\sum_{p\in\mathcal{P}(i)}
\log
\frac{\exp\!\big(s_{ip}/\tau\big)}
{\sum\limits_{a\neq i}\exp\!\big(s_{ia}/\tau\big)}
\right).
\]
To further tighten same-region clusters, we add a batchwise intra-class variance term.
Let $\mathcal{R}_\mathcal{B}$ be the set of region labels present in the batch and $\mathcal{B}_r=\{i\in\{1,\dots,N\}:y_i=r\}$, with class mean
$\mu_r=\frac{1}{|\mathcal{B}_r|}\sum_{i\in\mathcal{B}_r} z_i$.
The variance penalty is defined as:
\[
\mathcal{L}_{\text{var}}
=\frac{1}{|\mathcal{R}_\mathcal{B}|}\sum_{r\in\mathcal{R}_\mathcal{B}}
\frac{1}{|\mathcal{B}_r|}
\sum_{i\in\mathcal{B}_r}\!\left\|z_i-\mu_r\right\|_2^{2}.
\]
Our final objective is
\[
\mathcal{L}
=\mathcal{L}_{\text{sup}}+\lambda_{\text{var}}\,\mathcal{L}_{\text{var}},
\]
with $\lambda_{\text{var}}>0$ ($\lambda_{\text{var}}=0.05$ in our case) a tuning coefficient. No projection head and no view augmentations are used.

During training, inputs are sampled from either a single session or across sessions and subjects. In the single session setting, input pairs are not time-synchronized. In the multi-subject and session setting, pairs may originate from different individuals, sessions, and time points. This randomized, region-consistent sampling strategy ensures that the encoder learns region-specific neural signatures that are robust to subject and session variability.
\paragraph{Details.}
All model details and training parameters are provided in Appendix~\ref{app:cnn-encoder}, while training resources and times can be found in Appendix~\ref{A_resources}.

\subsection{Functional Transformer: Masked-Region Reconstruction}
To directly test our hypothesis that functional data-driven embeddings capture circuit-level relationships usable across subjects, we adopt masked-region reconstruction: withhold all channels from one region and require the model to predict them from the remaining regions. This task is label-free, runs on any multi-region dataset, isolates purely neural-circuit information, and naturally serves as a self-supervised pretraining objective for downstream decoding. For each window, multichannel LFP $\mathbf X\in\mathbb{R}^{C\times T}$ is split into \emph{sources} $\mathcal S$ (unmasked) and \emph{targets} $\mathcal T$ (masked) channels, and each channel has a 32-D functional embedding extracted once from a 10\,s snippet.

\textbf{Tokenization.}
A 1D conv tokenizer converts each source channel into $P$ time-patch features (width $d$). The channel’s functional embedding is mapped to width $d$, broadcast across its patches, concatenated with the conv features, reduced back to $d$, and augmented with a time-only positional code to form \emph{source tokens}. Targets have no signal; we build \emph{query tokens} by fusing a learned per-patch query base with the target’s functional embedding and the same positional code. Variable numbers of sources/targets are batched with padding masks.

\textbf{Architecture.}
A standard pre-LN encoder–decoder Transformer operates on these sequences: the encoder applies self-attention over source tokens to produce a memory; the decoder applies self-attention over queries followed by cross-attention to the memory, then a linear head maps each decoder token to a waveform patch that we concatenate to form the reconstruction. No subject IDs, subject-specific heads, or per-subject fine-tuning are used; a single shared model is trained across all subjects.

\textbf{Objective.}
We optimize a mixed loss that combines sample-wise MSE with a correlation term to emphasize waveform shape:
\[
\mathcal{L}=\mathrm{MSE}(\widehat{\mathbf Y},\mathbf Y)
\;+\;\lambda\big(1-\rho(\widehat{\mathbf Y},\mathbf Y)\big),\quad \lambda=0.05,
\]
where $\rho$ is mean Pearson correlation across targets. The correlation term counteracts the tendency of pure MSE to favor amplitude shrinkage/flat predictions under scale or SNR mismatch.

\textbf{Details.}
All token shapes, attention formulas (MHSA/MHA with Q/K/V sources), and masking/batching mechanics, model details and training parameters are provided in Appendix \ref{A_transformer}. Training resources and times can be found in Appendix~\ref{A_resources}

\section{Results}

\subsection{Functional embeddings reveal neural signature of each brain region across subjects (simulation study)} 

\begin{figure}[t]\center{\includegraphics[width=1\textwidth] {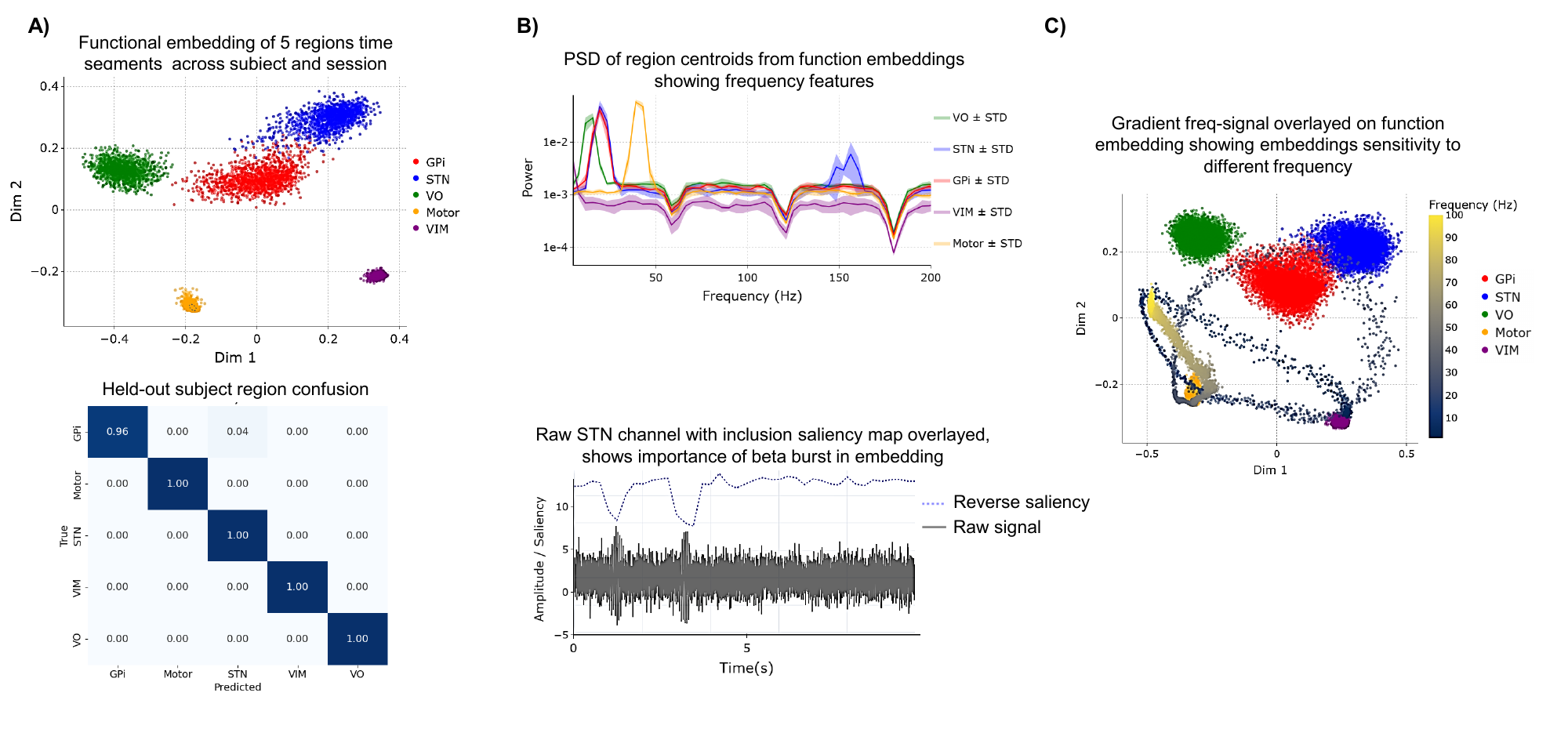}}
	\caption{\textit{Simulation-based validation of the functional embedding.}
        \textbf{(A)} Contrastive encoder trained on simulated data clusters test segments by region and generalizes to held-out subjects (scatter, top) with high accuracy (confusion matrix, bottom).
        \textbf{(B)} Top: Power spectra of signals nearest each region centroid match the ground-truth signatures; Bottom: Perturbation saliency shows largest embedding shifts when disrupting signature bursts.
        \textbf{(C)} Smooth frequency sweeps trace continuous trajectories in the embedding, passing near region centroids tuned to corresponding bands, indicating a locality-sensitive functional map.
        \vspace{-4mm}}
	\label{fig2}
\end{figure}

To assess whether the functional embedding network captures physiologically meaningful neural signatures, we designed a series of interpretability analyses using both simulated and analytic approaches.\\
\textbf{Simulation-based validation with ground-truth neural signatures.}
We first generated a simulated dataset in which each brain region was assigned a parameterized neural signature, primarily defined in the spectral domain, and recordings were synthesized across 10 subjects and 2 sessions (additional details in Appendix~\ref{A_simulation}). Subject/session variability was introduced by scaling, frequency shifts, and noise (Fig.~\ref{A_fig_simulation}). Training on a subset of subjects and evaluating on held-out subjects, the functional embedding network trained with contrastive learning (PSC) correctly clustered recordings by region with high accuracy (Fig.~\ref{fig2}A). This demonstrates that the embedding framework is capable of extracting stable, region-specific neural signatures even in the presence of cross-subject variability.\\
\textbf{Perturbation-based attribution.}
We next probed which input features (sub-windows) contributed most strongly to embedding location. Short temporal segments of the input were selectively frozen while other parts of the signal were perturbed, and the resulting displacement in embedding space was quantified (saliency map). Freezing segments overlapping the bursts that defined the region’s neural signature produced the smallest embedding shifts (Fig.~\ref{fig2}B bottom). This confirms that the embeddings are sensitive to the physiologically relevant components of the signal.\\
\textbf{Spectral properties of embedding centroids.}
To further link the embedding space to interpretable features, we computed centroids for each region cluster and examined their spectral properties. The power spectral densities of signals nearest to each centroid revealed frequency characteristics that matched the region’s simulated neural signature (Fig.~\ref{fig2}B top). This shows that the embedding clusters correspond to distinct and physiologically interpretable frequency-domain patterns.\\
\textbf{Smoothness of the functional map.} 
Finally, we evaluated the continuity of the embedding space in encoding signal properties, by presenting pure sinusoidal inputs with gradually increasing frequency. The embeddings formed smooth trajectories through the functional space, passing near centroids of regions tuned to corresponding frequency bands (Fig.~\ref{fig2}C). This demonstrates that the functional embedding defines a continuous map where novel signals can be meaningfully placed relative to known regions.
\subsection{Functional Embedding learned within individual subjects provide a locality sensitive embedding space encoding function}

\begin{figure}[t!]\center{\includegraphics[width=1\textwidth,clip=[] {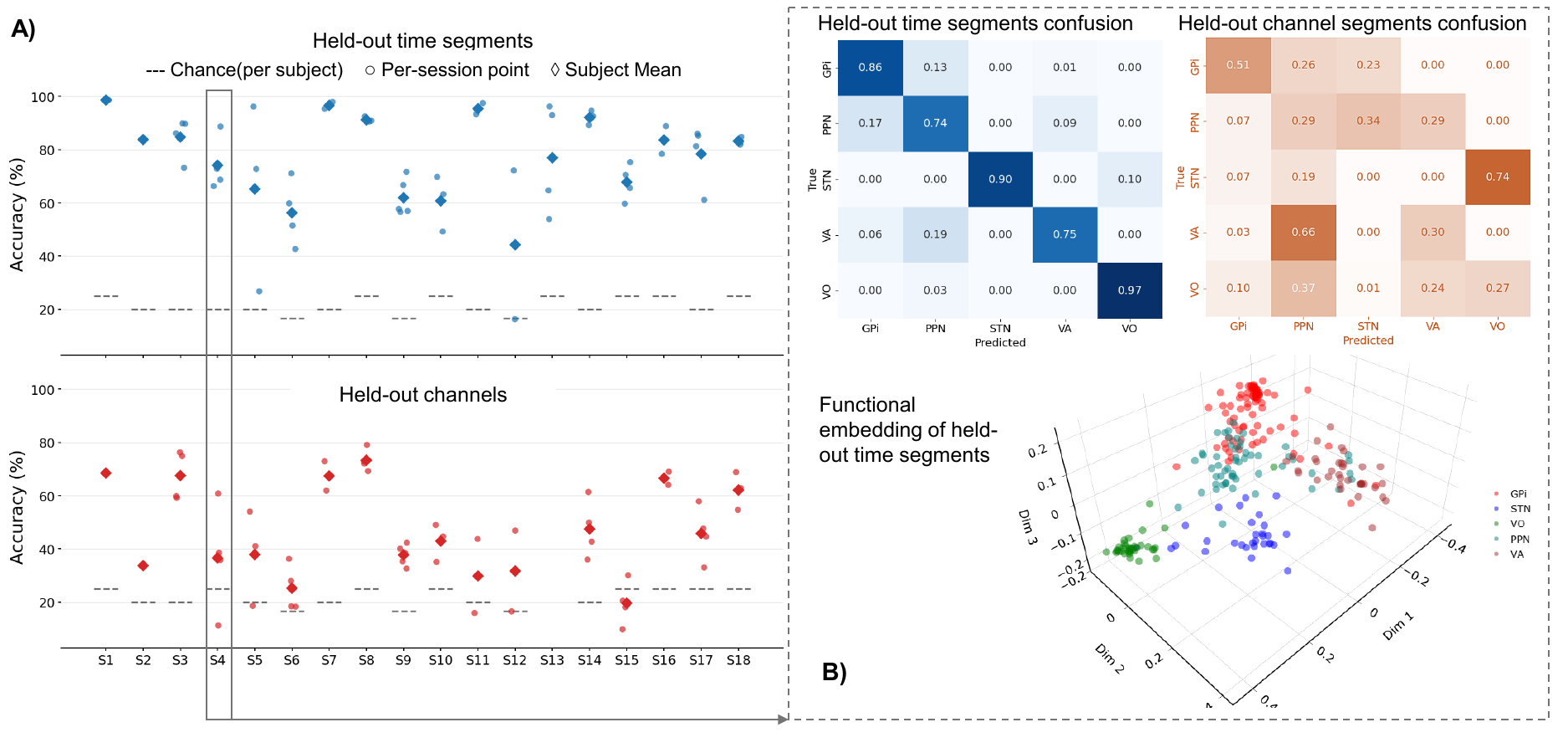}}
    \caption{\textit{Single-subject functional embeddings.}
    \textbf{(A)} Per-subject accuracies: top—held-out time segments from seen channels; bottom—held-out channels (only regions with \(>3\) channels). Diamonds denote subject means across sessions; circles denote session-level points; dashed lines indicate per-subject chance levels.
    \textbf{(B)} Example subject (S4): top — confusion matrices from \(k\)-NN on embeddings for held-out time segments (left) and held-out channels (right); bottom — 3D PCA visualization of the 32-D embedding space, color-coded by brain region.
    \vspace{-6mm}}
	\label{fig3}
\end{figure}

Across 20 subjects (most with multiple sessions), per-subject functional encoders (using PSC) trained on 10~s segments achieved a mean test accuracy of \(75.78\% \pm 17.90\%\) (M\(\pm\)SD) on held-out time segments from electrodes seen during training (Fig.~\ref{fig3}A, top). Under the stricter held-out channel evaluation (regions with \(>3\) channels), accuracy remained above chance at \(45.79\% \pm 18.44\%\) across subjects, with some reaching \(\sim70\%\) (Fig.~\ref{fig3}A, bottom). For an example subject (S4), the 32-D embeddings projected via PCA exhibit clear region-specific separation for held-out time segments, reflecting a locality-preserving organization of the learned space (Fig.~\ref{fig3}B, bottom). Applying \(k\)-NN on these embeddings yields confusion matrices that are strongly diagonal for held-out time segments and noticeably weaker for held-out channels in this subject (Fig.~\ref{fig3}B, top).
Overall, subject-specific functional encoders demonstrate reliable within-subject, region-sensitive representations; channel-level generalization is lower than time-segment generalization, likely due to the limited number of electrodes per region within a subject (\(\sim6\text{--}12\)), yet on average remains above chance.

\subsection{Aggregated data over subjects encode brain regions}

\begin{wrapfigure}{r!}{0.5\textwidth}
\includegraphics[width=0.5\textwidth]{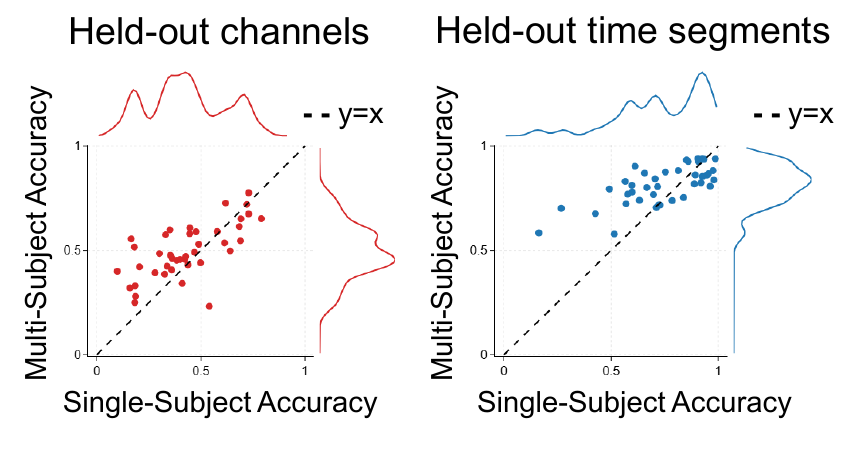}
	\caption{\textit{Cross-subject aggregation} Head-to-head comparison of the multi-subject model (y-axis) versus single-subject models (x-axis) for held-out time (left) and held-out channels (right). Each point is a recording session; the dashed line denotes \(y=x\). Marginal densities summarize performance distributions.
    \vspace{-5mm}}
	\label{fig4a}
\end{wrapfigure}

\textbf{Functional encoder successfully learns combined dataset.}
When the functional embedding was trained jointly on the entire dataset—sampling similar/dissimilar pairs across subjects and sessions (using PSC), it learned a global, subject-independent mapping with region-specific neural signatures.
On unseen time segments, it achieved \(80.71\% \pm 11.41\%\) (M\(\pm\)SD) accuracy. The same network generalized zero-shot to held-out channels, yielding \(49.18\% \pm 12.11\%\) accuracy (above per-subject chance). Compared to training a separate CNN per subject, the joint model performed better by \(\sim 5\%\) on average for both held-out time segments and held-out channels without any subject-specific fine-tuning, demonstrating successful aggregation across 20 subjects in a unified model.
A head-to-head comparison per recording session shows the multi-subject model surpasses single-subject models in most cases, especially where single-subject performance was low, likely due to limited channels per region (Fig.~\ref{fig4a}). The performance variability across subjects also decreased under the multi-subject model, consistent with a more robust and generalizable encoder. \\ \\ \\

\begin{figure}[t]\center{\includegraphics[width=1\textwidth,clip=false] {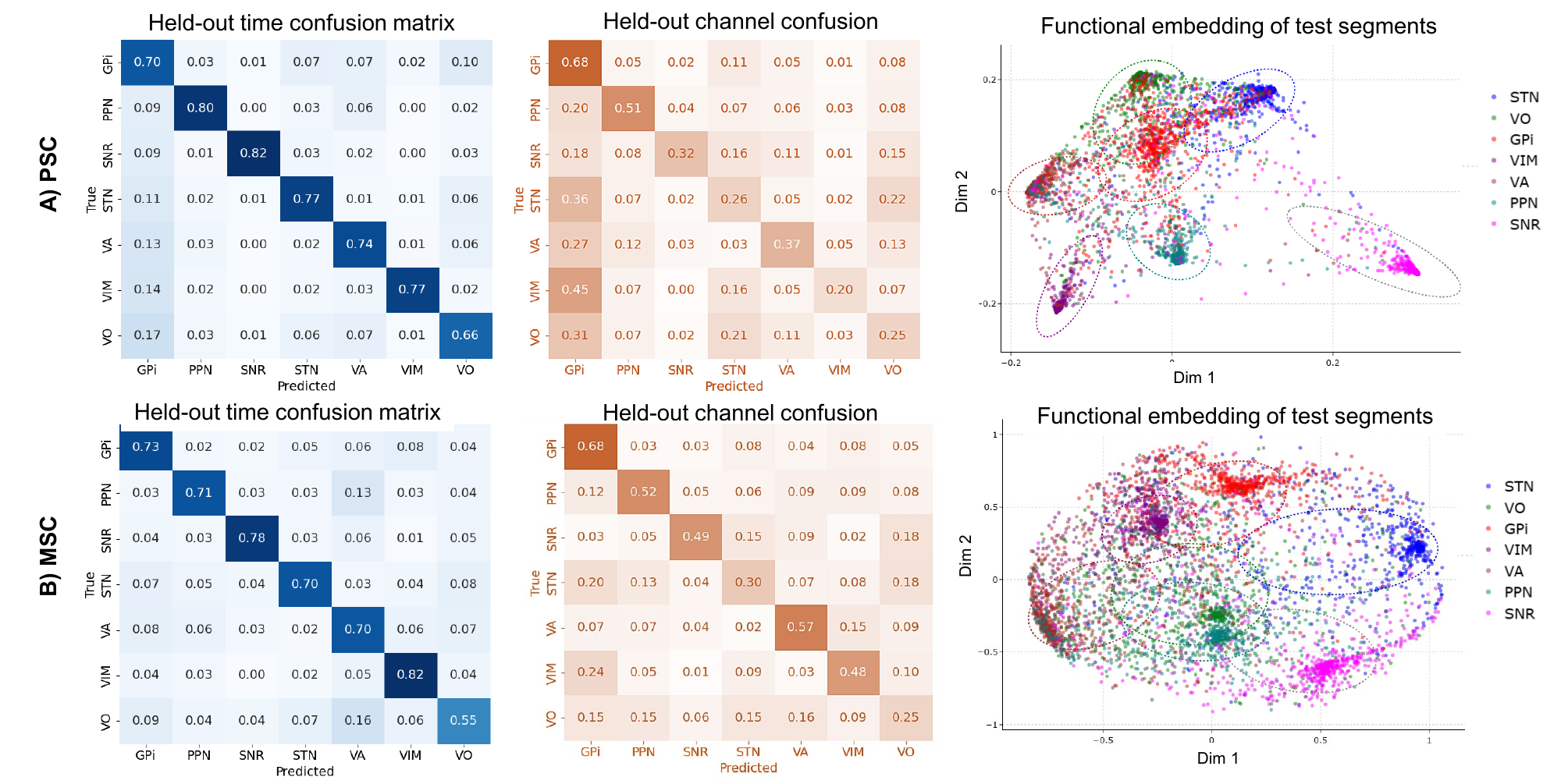}}
	\caption{
		\textit{Comparison of contrastive Methods (PSC vs. MSC)}
        \textbf{A) PSC:} Row-normalized confusion matrices from \(k\)-NN classification in the learned space for held-out time (left) and held-out channels (center), plus a 2D PCA view of test-segment embeddings (right).
        \textbf{B) MSC:} The same evaluations for MSC. Embeddings are organized on a hypersphere with broader, angle-separated region clouds. Dotted ellipses highlight region-consistent structure in both methods.
        \vspace{-3mm}}
	\label{fig4b}
\end{figure}

\textbf{Contrastive objective comparison.}
We compared two previously mentioned contrastive methods trained under the same multi-subject regime (Fig.~\ref{fig4b}). Both PSC and MSC created embedding spaces that segregates brain regions and clusters samples from same region close together. PSC yields slightly higher accuracy on held-out time segments, whereas MSC clearly outperforms on held-out channels, indicating stronger channel-robust generalization (\ref{fig4b}). The embedding geometries differ: MSC computes dot products between \(\ell_2\)-normalized vectors (cosine similarity), which constrains representations to a hypersphere in the 32-D space and emphasizes angular separation; PSC operates in Euclidean feature space, producing more compact, centroid-like clusters. Both embeddings are tested as tokenizers for the transformer.
\subsection{Functional embeddings enable aggregation of neural data over subjects using transformers}

\begin{figure}[h!]\center{\includegraphics[width=1\textwidth,clip=false] {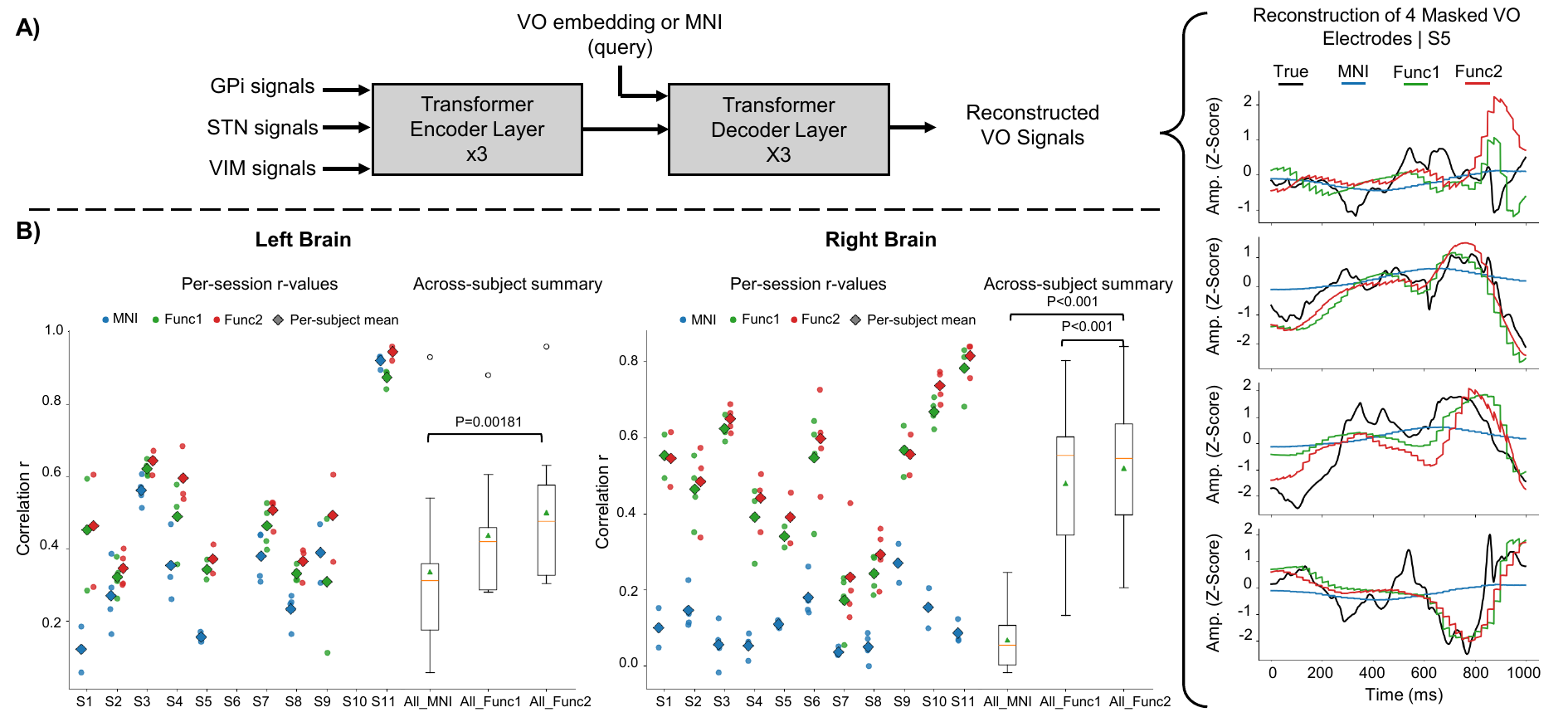}}
	\caption{
        \textit{Ablation of coordinate systems used in masked–region reconstruction across subjects.}
        \textbf{(A)} Schematic: GPi/STN/VIM signals enter a shared encoder as source tokens; VO channels are queried via either MNI or functional embeddings, and the decoder reconstructs VO waveforms.
        Example of 4 synchronous VO channels reconstruction attempts with 3 different coordinate systems are provided on the right side. \textbf{(B)} Per–subject results (S1–S11): per–session points and subject means show consistent gains for Functional–1/2 over MNI for both brain sides.
        Across–subject box plots (All MNI / All Func1 / All Func2) summarize higher correlation $r$ for functional embeddings; Functional–2 is significantly better than MNI.
        All models use a single shared trunk with no subject IDs, subject–specific heads, or fine–tuning. }
      \label{fig:functional_embeddings_recon}
	\label{fig5}
\end{figure}

We tested whether functional embeddings provide a subject-agnostic coordinate system that improves cross-subject pooling for masked-region reconstruction. In each evaluation window, all channels from a target region (here, VO) were withheld, signals from the remaining regions (i.e., GPi, STN, and VIM) were encoded as source tokens, and the model queried the withheld VO channels using their coordinate system. An ablation study was done on the type of the coordinate system used, namely, functional embeddings or MNI coordinates (Fig.~\ref{fig5}A). We compared three otherwise-identical transformers that differed only in the channel embedding (coordinate system) used at tokenization: MNI coordinates, Functional-1 (PSC), and Functional-2 (MSC). Performance was summarized by the Pearson correlation ($r$) between predicted and true waveforms, computed per target channel and window, then aggregated per subject using Fisher-z averaging.
Importantly, across subjects, functional embeddings significantly improved reconstruction accuracy relative to MNI (purely anatomical) for both hemispheres (Fig.~\ref{fig5}B). Per-subject dot plots (Fig.~\ref{fig5}B) show consistent upward shifts from MNI to Functional-1 and Functional-2. Over all subjects (box plots fig.~\ref{fig5}B), Functional-2 outperformed MNI with statistical significance (paired test on subject-level $r$; $p\approx0.002$ in this cohort), while Functional-1 showed a positive trend that was smaller and not reliably significant. Importantly, all results were obtained without subject-ID tokens, subject-specific heads, or fine-tuning, i.e., a single shared model across all subjects.
Qualitatively, functional embeddings also produced more differentiated reconstructions across simultaneously recorded VO channels (Fig.~\ref{fig5}B). Because the four VO electrodes often shared nearly identical or uncertain MNI coordinates, the MNI-based model tended to generate similar waveforms for targets. In contrast, Functional-1/2 leveraged each channel’s functional identity, yielding channel-specific predictions that better matched the heterogeneous ground-truth VO activity. We benchmarked our model against subject-specific baselines trained separately for each individual, including a linear causal multi-output FIR, a Temporal Convolutional Network (TCN), a 2-layer GRU, a CopyBest correlation baseline, and a Zero predictor (details in Appendix~\ref{A_baseline}). Across both left and right hemispheres, the \textbf{Transformer + Functional Embedding-2 (FUNC2)} trained on pooled multi-subject data achieved the highest reconstruction accuracy, statistically significantly outperforming all per-subject baselines (paired two-sided $t$-tests with Holm and BH/FDR correction; all corrected $p<0.001$). These results indicate that, under a unified functional coordinate system, the proposed transformer effectively aggregates information across subjects to learn stronger inter-areal relationships for signal reconstruction.

\vspace{-1mm}
\section{Discussion}
We asked whether neural signals can be aligned across subjects by \emph{function} rather than standard coordinates, and whether such a representation can support scalable transformer models operating over variable and sparse electrode sets. Our findings support both claims. A contrastively trained encoder learns a subject-agnostic \emph{functional identity} for each channel: short segments from the same deep-brain region cluster in a locality-sensitive latent space and generalize across time, channels, and subjects. Conditioning transformers on these functional tokens enables cross-subject aggregation without subject IDs or subject-specific heads, yielding improved masked-region reconstruction over atlas-based (MNI) tokenization. Together, this suggests that data-derived coordinates provide a more reliable backbone for heterogeneous intracranial datasets than static anatomical alignment.

Two recent directions highlight this distinction. \citet{mentzelopoulos2024seegnificant} mix electrode signals with MNI coordinates and finish with subject-specific heads. Our approach removes both subject identifiers and atlas coordinates, replacing them with a learned functional identity. This matters when localization is uncertain or overlapping; functional tokens consistently yield clearer reconstructions than MNI tokens under otherwise identical transformers, consistent with their own finding that positional encodings did not improve performance. \citet{chau2025popT} propose Population Transformer (PopT), which aggregates frozen single-channel embeddings with positional encodings. Our pipeline is complementary: we pretrain the \emph{channel identity} itself via contrastive learning, producing region-consistent functional coordinates that can serve as tokens for any aggregator. Conceptually, PopT learns the population-level \emph{aggregator}, while our method learns the population-level \emph{coordinate system}.

A functional coordinate system enables (i) subject-agnostic pooling across irregular montages, (ii) plug-and-play tokenization for large transformer backbones, and (iii) reusable pretraining across tasks such as decoding, forecasting, and imputation. Because our recordings were collected under flexible rest/movement recording sessions rather than narrowly defined tasks, the approach extends naturally to larger, more varied datasets—the setting where transformers excel.

Future work should relax reliance on region labels during contrastive training, exploring self- and weakly supervised objectives that preserve functional locality at scale. Extending beyond basal ganglia-thalamic circuits to cortical ECoG and spikes will test generality. Finally, a fuller head-to-head comparison against population-level pretraining frameworks such as PopT remains important. Bridging the two directions—functional coordinates for channel identity and ensemble-level pretraining—may yield the best of both worlds.

\section*{Ethics Statement}
All human neural data analyzed in this study were collected under protocols approved by the Institutional Review Boards (IRBs) of the participating medical centers. Written informed consent was obtained from all participants or their legal guardians (for minors), with assent obtained from children when appropriate. Data handling complied with HIPAA and institutional privacy policies: recordings were de-identified prior to analysis, stored on secure, access-controlled systems, and used solely for research purposes. This work does not involve clinical decision-making, patient intervention beyond standard clinical care, or vulnerable-population recruitment beyond the clinical cohort already undergoing DBS evaluation. No attempt was made to re-identify participants. Due to ethical and regulatory constraints governing research with human subjects, the raw neural recordings cannot be publicly released; de-identified data may be shared upon reasonable request and execution of a data use agreement consistent with IRB approvals. The authors affirm compliance with the ICLR Code of Ethics and disclose no conflicts of interest or external sponsorship that could influence the work.

\section*{Reproducibility Statement}
All code used in this study is released to support reproducibility at
\url{https://github.com/ICLR-Functional-Embedding/ICLR2026_Functional_Map}. Complete and detailed explanation of datasets used, data preprocessing, training details and training resources and  are provided in Appendix.

\section{Acknowledgments}
This work was supported by funding from the Raynor Cerebellum Project.

\bibliography{iclr2026_conference}
\bibliographystyle{iclr2026_conference}

\appendix
\section{Appendix}

\renewcommand{\thefigure}{A.\arabic{figure}}
\renewcommand{\thetable}{A.\arabic{table}}
\renewcommand{\theequation}{A.\arabic{equation}}
\setcounter{figure}{0}
\setcounter{table}{0}
\setcounter{equation}{0}

\subsection{Dataset and Clinical Metadata}
\label{A_dataset}
\paragraph{Subjects.}
Twenty (20) subjects who underwent DBS surgery for the treatment of dystonia participated in this study (Table \ref{table1}). The diagnosis of dystonia was established by a pediatric movement disorder specialist (T.D.S.) using standard criteria \citep{sanger2003classification}. All patients provided signed informed consent for surgical procedures in accordance with standard hospital practice at \iftrue  Children’s Health Orange County (CHOC) or Children’s Hospital Los Angeles (CHLA)\fi. The study protocol was approved by the institutional review boards (IRBs) of both hospitals. The patients, or parents of minor patients, also signed an informed consent to the research use of electrophysiological data and provided Health Insurance Portability and Accountability Act (HIPAA) authorization for the research use of protected health information. 

\begin{table}[t]
\caption{\textit{Demographic characteristics:} 20 male subjects, between 5 and 24 years old. H-ABC: Hypomyelination with Atrophy of the Basal Ganglia and Cerebellum; CP: Cerebral palsy; GA1: Glutaric aciduria type 1; MEPAN: Mitochondrial Enoyl CoA Reductase Protein-Associated Neurodegeneration; KMT2B: K-specific Methyltransferase 2B; PKAN: Pantothenate Kinase-Associated Neurodegeneration; MYH2: Myosin Heavy Chain 2; GPi: globus pallidus internus; VoaVop: ventral oralis anterior-posterior; STN: subthalamic nucleus; VA: ventral anterior; VPL: ventral posterolateral; VIM: ventral intermediate; PPN: Pedunculopontine nucleus; SNR: Substantia nigra pars reticulata; CM: Centromedian nucleus; CL: Centrolateral nucleus. \\}\label{table1}

\centering
\begin{tabular}{ccc}
\hline 
\textbf{Subject} & \textbf{Etiology} &  \textbf{Implanted Regions}\\
\hline \\
1   & Tourette & NA, VoaVop, VIM, STN, CMP, GPi\\
2  & CP & VA, VoaVop, VIM, STN, GPi\\
3   & GA1 & VA, VoaVop, VIM, STN, GPi\\
4   & MEPAN & VA, VoaVop, VIM, PPN, STN, GPi\\
5   & MEPAN & VA, VoaVop, VIM, PPN, STN, GPi\\
6   & Unknown  & VA, VoaVop, PPN, STN, GPi \\
7   & KMT2B  & VA, VoaVop, VIM, PPN, STN, GPi\\
8   & CP  & VA, VoaVop, VIM, PPN, STN, GPi\\
9   & GA1 & VA, VoaVop,VIM, PPN, STN, GPi\\  
10   & PKAN  & VoaVop, VIM, PPN, SNR, STN, GPi\\
11  & CP & VA, VoaVop, VIM, PPN, SNR, STN, GPi\\
12   & MYH2 & VoaVop, VIM, PPN, SNR, STN, GPi\\
13   & CP  & VoaVop, VIM, PPN, SNR, STN, GPi\\
14   & GA1 & VoaVop, VIM, PPN, SNR, STN, GPi\\
15   & CP &  VoaVop, VIM, STN, GPi\\
16  & CP &  VA, VoaVop, VIM, PPN, STN, GPi\\
17  & KMT2B & VoaVop, VIM, CM, STN, GPi\\
18  & CP & VoaVop, VIM, CM, STN, GPi\\
19  & CP & VoaVop, VIM, CM, SNR, STN, GPi\\
20  & KMT2B & VoaVop, VIM, CM, CL, STN, GPi\\

\end{tabular}
\end{table} 

\paragraph{Experimental protocol and recording sessions.}
Participants performed self-paced finger-to-nose reaching or hand-squeeze movements, depending on motor ability. The clinical team administered short blocks consisting of \(\sim\)60\,s of active movement followed by \(\sim\)30\,s of rest, repeated at least four times per hand. Trajectory and speed were unconstrained, and block durations were adjusted as needed to accommodate each participant. For all analyses, recordings were not segmented by trials or active/rest windows; instead, we used the entire continuous LFP recordings from each session. This yields behaviorally diverse, non-stationary signals, underscoring the advantage of our representation-learning approach, which remains robust to noise and variability in naturalistic clinical settings.

\paragraph{Recording electrodes and spatial coverage.}
\label{app:electrode}
Local field potentials (LFPs) were recorded from high-impedance microwire (``micro-contact'') electrodes on temporary Stereo-EEG (SEEG) depth leads (AdTech MM16C; AdTech Medical Instrument Corp., Oak Creek, WI, USA) implanted with standard stereotactic procedures \cite{liker2024stereotactic,liker2019pediatric}. 
Up to 12 SEEG leads were placed per subject to sample candidate DBS targets. 
Each lead includes six low-impedance ring macro-contacts (1–2~k\(\Omega\); 2~mm height) and ten high-impedance 50-\(\mu\)m micro-contacts (70–90~k\(\Omega\)). 
Micro-contact signals were sampled at 24{,}414~Hz using a PZ5M 256-channel digitizer and RZ2 processor, with data stored on an RS4 system (Tucker-Davis Technologies, Alachua, FL, USA). 
All analyses in this paper use LFPs from the micro-contacts; macro-contact recordings were not used.

Table~\ref{A_table_electrodes} summarizes electrode spatial coverage per subject (S1–S20), listing the number of unique recording channels in each brain region alongside the total recording time and electrode-hours. Brain‐region assignments were performed by an expert via manual review of co-registered MRI–CT volumes. Recording time is computed as the sum of session lengths for each subject, while electrode-hours equal each subject’s total recording time multiplied by their total channel count. Across all subjects, the dataset comprises 9.96 recording hours and 442.86 electrode-hours. 

\begin{table}
\caption{\textit{Recording coverage by subject and brain region:} “Recording time” sums recording lengths over all sessions for each subject. “Electrode hours” multiply each subject’s total recording time by that subject’s number of recorded channels across regions.\\}
\label{A_table_electrodes}
\centering
\footnotesize
\setlength{\tabcolsep}{4pt}
\begin{tabular}{cccccccccc}
\textbf{Subject (sessions)} & \textbf{GPi} & \textbf{PPN} & \textbf{SNR} & \textbf{STN} & \textbf{VA} & \textbf{VIM} & \textbf{VO} & \textbf{Recording time (hrs)} & \textbf{Electrode hours} \\
\hline \\
S1 (2 sessions)  & 19 & 0  & 0  & 6  & 0  & 9  & 6  & 1.60 & 63.81 \\
S2 (3 sessions)  & 12 & 0  & 0  & 6  & 6  & 6  & 6  & 0.36 & 12.97 \\
S3 (3 sessions)  & 12 & 0  & 0  & 6  & 6  & 6  & 6  & 0.37 & 13.40 \\
S4 (2 sessions)  & 18 & 12 & 0  & 6  & 8  & 0  & 8  & 0.23 & 12.06 \\
S5 (4 sessions)  & 18 & 6  & 0  & 6  & 8  & 0  & 8  & 0.62 & 28.36 \\
S6 (4 sessions)  & 8  & 3  & 0  & 6  & 8  & 0  & 6  & 0.47 & 14.46 \\
S7 (3 sessions)  & 12 & 6  & 0  & 6  & 12 & 0  & 6  & 0.53 & 22.29 \\
S8 (4 sessions)  & 12 & 6  & 0  & 6  & 6  & 6  & 6  & 0.46 & 19.42 \\
S9 (2 sessions)  & 21 & 6  & 0  & 8  & 11 & 0  & 4  & 0.24 & 11.94 \\
S10 (4 sessions) & 29 & 9  & 0  & 10 & 0  & 0  & 4  & 0.44 & 23.07 \\
S11 (5 sessions) & 14 & 9  & 14 & 6  & 15 & 0  & 8  & 0.53 & 35.24 \\
S12 (3 sessions) & 25 & 9  & 0  & 10 & 0  & 0  & 4  & 0.30 & 14.64 \\
S13 (2 sessions) & 8  & 9  & 12 & 4  & 0  & 0  & 4  & 0.55 & 20.27 \\
S14 (2 sessions) & 15 & 8  & 6  & 6  & 6  & 0  & 6  & 0.25 & 11.74 \\
S15 (4 sessions) & 19 & 0  & 0  & 8  & 0  & 9  & 2  & 0.61 & 23.27 \\
S16 (4 sessions) & 15 & 10 & 0  & 6  & 8  & 0  & 8  & 0.63 & 29.51 \\
S17 (4 sessions) & 18 & 0  & 0  & 10 & 0  & 8  & 4  & 0.50 & 20.18 \\
S18 (2 sessions) & 20 & 0  & 0  & 11 & 0  & 10 & 6  & 0.25 & 11.84 \\
S19 (4 sessions) & 26 & 12 & 9  & 9  & 0  & 0  & 2  & 0.51 & 29.80 \\
S20 (3 sessions) & 20 & 0  & 0  & 12 & 0  & 10 & 8  & 0.49 & 24.62 \\
\hline
\textbf{Total} & \textbf{341} & \textbf{105} & \textbf{41} & \textbf{148} & \textbf{94} & \textbf{64} & \textbf{112} & \textbf{9.96} & \textbf{442.86} \\
\end{tabular}
\end{table}

\begin{figure}[h!]\center{\includegraphics[width=1\textwidth,clip=false] {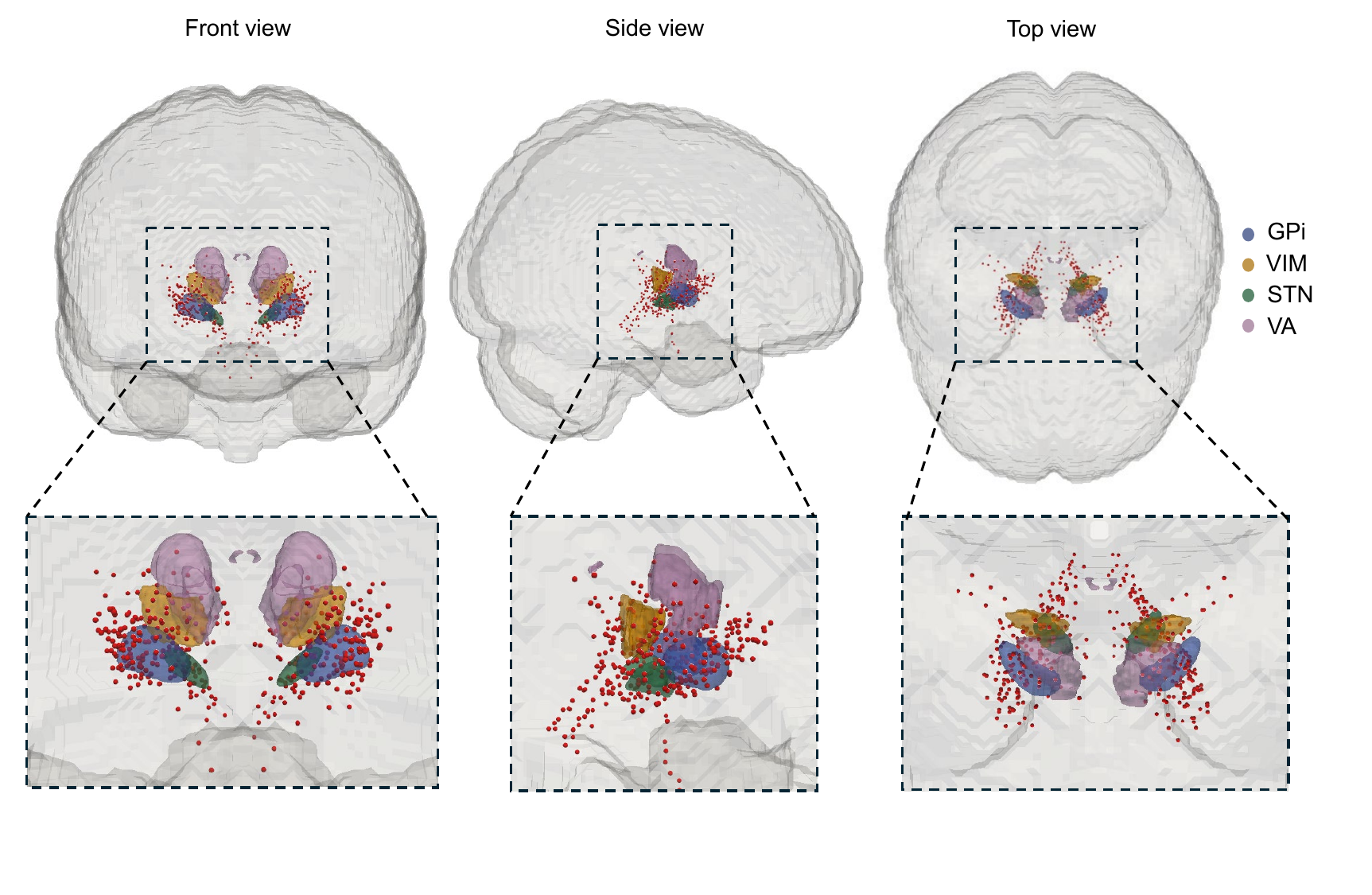}}
	\caption{
        \textit{3 views of aggregated micro-contacts locations over 11 subjects relative to 4 deep brain regions.}}
	\label{A_figure_electrodes}
\end{figure}

\paragraph{MNI coordinate extraction.}
MNI coordinates for a total of 11 subjects yielding approximately 600 macro-contact were extracted. Structural T1-weighted (T1W) MRI scans for all 11 subjects were nonlinearly normalized to MNI space using the Advanced Normalization Tools (ANTs) framework \citep{avants2008symmetric}, a widely adopted tool for high-accuracy nonlinear image registration \citep{tustison2019longitudinal,bartel2019non}.
Postoperative CT scans were rigidly co-registered to each subject’s preoperative T1W MRI using
the FLIRT (FMRIB’s Linear Image Registration Tool) module of the FMRIB Software Library
(FSL). The nonlinear warping transformations obtained from ANTs normalization of the T1W
images were subsequently applied to the co-registered CT scans, thereby aligning them to MNI
space.
Following spatial normalization, approximately 600 macro-contacts were localized across all
patients. Contact identification was performed using histogram analysis of CT metal artifacts,
with the voxel exhibiting the maximum CT intensity within each artifact region selected as the
contact center coordinate. These standardized MNI coordinates of macro-contacts were then used to generate known relative positions for micro-contacts which recorded the LFP signals. micro-contact locations were then used as inputs for the machine learning models. Figures \ref{A_figure_electrodes} shows all the micro-electrode locations from 11 subjects relative to 4 deep brain regions.  

\subsection{Additional Signal Preprocessing Details}
\label{A_preprocessing}
All preprocessing was performed per channel with zero-phase filtering. We describe the common steps followed by branch-specific steps for the functional encoder and transformer experiments.

\paragraph{Common steps.}
\begin{enumerate}
    \item \textbf{Anti-aliasing and resampling.} A 4th-order Butterworth low-pass filter with cutoff 500~Hz was applied, then signals were resampled to 1{,}000~Hz .
    \item \textbf{Line-noise removal.} Narrowband notches were applied at 60, 120, and 180~Hz using an IIR notch filter with quality factor $Q{=}30$.
    \item \textbf{Session-wise normalization.} For each session, the signal was standardized by subtracting the session mean and dividing by the session standard deviation.
\end{enumerate}

\paragraph{Functional encoder (contrastive) pipeline.}
\begin{enumerate}
    \item \textbf{Windowing.} Continuous recordings were segmented into non-overlapping 10~s windows.
    \item \textbf{Robust artifact rejection.} Let $x$ denote the full-session signal values for a channel. Define
    \[
        \mathrm{med} = \mathrm{median}(x), \quad
        \widehat{\sigma} = 1.4826 \cdot \mathrm{median}\!\left(|x - \mathrm{med}|\right) \;\; (\text{MAD-based robust std}).
    \]
    Windows were discarded if any sample fell outside $[\mathrm{med} - \tau \widehat{\sigma}, \; \mathrm{med} + \tau \widehat{\sigma}]$, with threshold $\tau=10$ in our implementation.
    \item \textbf{Splits.} Non-overlapping 10\,s windows were assigned to splits without shuffling in chronological order: the first \(70\%\) of windows to training, the next \(15\%\) to validation, and the final \(15\%\) to test. Entire windows were assigned to a single split (no window was divided across sets).

\end{enumerate}

\paragraph{Transformer pipeline.}
\begin{enumerate}
    \item \textbf{Low-pass filtering.} After the notch filters, a 2nd-order Butterworth low-pass filter with 50~Hz cutoff was applied.
    \item \textbf{Chronological splits.} The first $70\%$ of each session (in time) was used to construct training examples; the next $15\%$ and final $15\%$ were used for validation and test, respectively.
\end{enumerate}

\subsection{Additional details on simulation study}
\label{A_simulation}
To evaluate our embedding methods in a fully controlled setting, we constructed a synthetic dataset designed to reflect known region-specific neural signatures while allowing precise control over subject-, session-, and electrode-level variability.

The dataset includes:
\begin{itemize}
    \item $N = 10$ subjects
    \item $S = 2$ sessions per subject
    \item $R = 5$ brain regions per session: GPi, STN, VO, Motor Cortex, and VIM
    \item $C = 4$ electrodes (channels) per region
    \item $f_s = 1000$~Hz sampling rate
\end{itemize}

Each recording channel was synthesized by combining a region-specific neural signal generator with additive Gaussian noise, modulated by three hierarchical levels of parameters: subject-level, session-level, and electrode-level.

\textbf{1. Region-Specific Neural Signatures}

Each brain region was associated with one or more characteristic oscillatory components:

\begin{table}[h]
\centering
\begin{tabular}{ll}

\textbf{Region} & \textbf{Signature Components} \\
\hline 
GPi   & Beta bursts ($\sim$20 Hz) \\
STN   & Beta bursts + HFOs ($\sim$150 Hz) \\
VO    & Spindle bursts ($\sim$12--16 Hz) \\
Motor & Continuous gamma activity ($\sim$40 Hz) \\
VIM   & Slow oscillations ($\sim$1 Hz) \\

\end{tabular}
\end{table}

Each oscillatory component was implemented using sine waves modulated by windowed envelopes (e.g., Hann windows for bursts), with frequency jitter and amplitude variation across subjects and channels.

\textbf{2. Parameter Hierarchy}

We define the following hierarchical parameter sets:

\paragraph{Subject-Level Parameters: $\boldsymbol{\theta_\text{subject}}$}
Sampled once per subject, controlling global oscillation strength and structure:
\[
\theta_\text{subject} = \{
\beta_\text{gain},\ \beta_\text{freq\_offset},\ \beta_\text{burst\_prob},\
\text{spindle}_\text{gain},\ \text{spindle}_\text{freq\_offset},\ \text{spindle}_\text{burst\_prob},\
\gamma_\text{gain},\ \gamma_\text{freq\_offset},  \]
\[
\text{HFO}_\text{gain},\ \text{HFO}_\text{freq\_offset},\ 
\text{slow}_\text{gain},\ \text{slow}_\text{freq\_offset},\ 
\text{noise}_\text{level}
\}
\]
Sampled from the following distributions:
\begin{align*}
\beta_\text{gain} &\sim \mathcal{U}(0.8,\ 1.5) &
\beta_\text{freq\_offset} &\sim \mathcal{U}(-2.0,\ 2.0) &
\beta_\text{burst\_prob} &\sim \mathcal{U}(0.3,\ 0.7) \\
\text{spindle}_\text{gain} &\sim \mathcal{U}(0.8,\ 1.3) &
\text{spindle}_\text{freq\_offset} &\sim \mathcal{U}(-1.0,\ 1.0) &
\text{spindle}_\text{burst\_prob} &\sim \mathcal{U}(0.2,\ 0.5) \\
\gamma_\text{gain} &\sim \mathcal{U}(0.3,\ 0.7) &
\gamma_\text{freq\_offset} &\sim \mathcal{U}(-5.0,\ 5.0) \\
\text{HFO}_\text{gain} &\sim \mathcal{U}(0.1,\ 0.3) &
\text{HFO}_\text{freq\_offset} &\sim \mathcal{U}(-10.0,\ 10.0) \\
\text{slow}_\text{gain} &\sim \mathcal{U}(0.8,\ 1.2) &
\text{slow}_\text{freq\_offset} &\sim \mathcal{U}(-0.3,\ 0.3) \\
\text{noise}_\text{level} &\sim \mathcal{U}(0.3,\ 0.4)
\end{align*}

\paragraph{Session-Level Parameters: $\boldsymbol{\theta_\text{session}}$}
Sampled per session:
\[
\theta_\text{session} = \{ \text{session\_noise} \sim \mathcal{U}(0.05,\ 0.15) \}
\]

\paragraph{Electrode-Level Parameters: $\boldsymbol{\theta_\text{electrode}}$}
Sampled per electrode:
\[
\theta_\text{electrode} = \{ \text{gain} \sim \mathcal{U}(0.9,\ 1.1),\ \text{electrode\_noise} \sim \mathcal{U}(0.03,\ 0.1) \}
\]

\textbf{3. Signal Model}

For a given region $r$, session $s$, and electrode $e$, the generated signal is defined as:
\[
\text{Signal}_{r,s,e}(t) = \text{Signature}_r(t;\ \theta_\text{subject},\ \theta_\text{electrode}) + \mathcal{N}\big(0,\ \sigma^2_\text{subject} + \sigma^2_\text{session} + \sigma^2_\text{electrode}\big)
\]

\begin{figure}[!ht]\center{\includegraphics[width=1\textwidth] {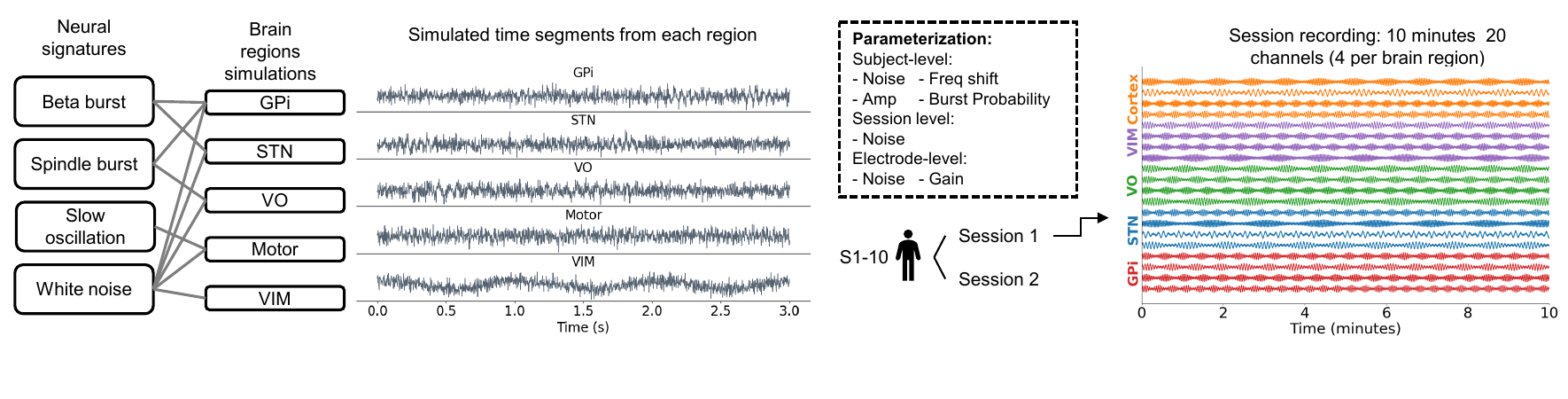}}
	\caption{\textit{Simulation dataset.} pipeline: each region is assigned a parameterized neural signature; recordings are generated across subjects/sessions with gain, frequency shifts, and noise parameters.}
	\label{A_fig_simulation}
\end{figure}

Each component of the signal is constructed from filtered oscillatory patterns with gain-scaled amplitudes, frequency offsets, and additive Gaussian noise aggregated from all three levels. 
The pipeline for generating simulated dataset is depicted in fig.~\ref{A_fig_simulation}. 

\subsection{CNN Functional Encoder: Architecture and Training}
\label{app:cnn-encoder}

\paragraph{Setup.}
We learn a subject-agnostic 32D embedding of 1D neural time series using convolutional neural network. 

\paragraph{Single-session encoder.}
The encoder $f_\theta$ is a 1D CNN with following blocks:
\[
\begin{aligned}
&\text{Conv1d}(1,64, k{=}15, s{=}4)\!\rightarrow\!\text{BN}(64)\!\rightarrow\!\text{GELU}\!\rightarrow\!\text{MaxPool}(3, s{=}2)\!\rightarrow\!\text{Dropout} \\
&\text{Conv1d}(64,64, k{=}11, s{=}2)\!\rightarrow\!\text{BN}(64)\!\rightarrow\!\text{GELU}\!\rightarrow\!\text{MaxPool}(3, s{=}2)\!\rightarrow\!\text{Dropout} \\
&\text{Conv1d}(64,64, k{=}7, s{=}2)\!\rightarrow\!\text{BN}(64)\!\rightarrow\!\text{GELU}\!\rightarrow\!\text{MaxPool}(3, s{=}2)\!\rightarrow\!\text{Dropout} \\
&\text{Conv1d}(64,64, k{=}5, s{=}2)\!\rightarrow\!\text{BN}(64)\!\rightarrow\!\text{GELU}\!\rightarrow\!\text{MaxPool}(3, s{=}2)\!\rightarrow\!\text{Dropout} \\
&\text{AdaptiveAvgPool1d}(10)\ \ \to\ \ \text{Flatten}\ (10\!\times\!64)\ \ \to\ \ \text{MLP: } \mathbb{R}^{640}\!\to\!\mathbb{R}^{32}\!\to\!\mathbb{R}^{32}.
\end{aligned}
\]
The output embedding is $\ell_2$-normalized.
This configuration has $117{,}504$ trainable parameters.

\textbf{Training.}
We train for $100$ epochs using AdamW \citep{LoshchilovHutter2019AdamW} (weight decay $10^{-4}$, $\beta_1{=}0.9$, $\beta_2{=}0.99$) with initial
learning rate $\mathrm{lr}{=}0.01$, batch size $128$, and a ReduceLROnPlateau scheduler (mode=\texttt{min}, patience=5,
factor=0.5, $\mathrm{lr}_{\min}{=}10^{-5}$). Dropout in conv blocks is $0.5$. The embedding head applies a
small dropout before the final linear layer and outputs unit-norm vectors.

\paragraph{Multi-subject / multi-session encoder}
For pooled training across subjects and sessions we use a deeper/wider encoder with the same design principles,
increasing channels and depth before the global pooling step:
\[
\begin{aligned}
&\text{Conv1d}(1,64, k{=}15, s{=}4)\!\rightarrow\!\text{BN}(64)\!\rightarrow\!\text{GELU}\!\rightarrow\!\text{MaxPool}(3, s{=}2)\!\rightarrow\!\text{Dropout} \\
&\text{Conv1d}(64,128, k{=}11, s{=}2)\!\rightarrow\!\text{BN}(128)\!\rightarrow\!\text{GELU}\!\rightarrow\!\text{MaxPool}(3, s{=}2)\!\rightarrow\!\text{Dropout} \\
&\text{Conv1d}(128,128, k{=}11, s{=}2)\!\rightarrow\!\text{BN}(128)\!\rightarrow\!\text{GELU}\!\rightarrow\!\text{MaxPool}(3, s{=}2)\!\rightarrow\!\text{Dropout} \\
&\text{Conv1d}(128,128, k{=}11, s{=}2)\!\rightarrow\!\text{BN}(128)\!\rightarrow\!\text{GELU}\!\rightarrow\!\text{MaxPool}(3, s{=}2)\!\rightarrow\!\text{Dropout} \\
&\text{Conv1d}(128,128, k{=}7, s{=}2)\!\rightarrow\!\text{BN}(128)\!\rightarrow\!\text{GELU}\!\rightarrow\!\text{Dropout} \\
&\text{Conv1d}(128,128, k{=}5, s{=}2)\!\rightarrow\!\text{BN}(128)\!\rightarrow\!\text{GELU}\!\rightarrow\!\text{Dropout} \\
&\text{AdaptiveAvgPool1d}(10)\ \ \to\ \ \text{Flatten}\ (10\!\times\!128)\ \ \to\ \ \text{MLP: } \mathbb{R}^{1280}\!\to\!\mathbb{R}^{32}\!\to\!\mathbb{R}^{32}.
\end{aligned}
\]
The output embedding remains $F{=}32$ and is $\ell_2$-normalized. This configuration has $694{,}464$ trainable parameters.
Dropout in conv blocks is set to $0.3$.

\textbf{Training.}
We use the same optimizer family and regularization as above (AdamW with weight decay $10^{-4}$, $\beta_1{=}0.9$,
$\beta_2{=}0.99$), increasing the batch size to $1024$ to leverage cross-subject aggregation. Learning-rate scheduling
follows the same ReduceLROnPlateau policy; the initial learning rate matches the single-session setup unless otherwise
stated. The embedding head outputs unit-norm vectors used by the contrastive criteria in the main text.

\subsection{Details of the functional transformer}
\label{A_transformer}

The functional embedding of each electrode serves as a subject-agnostic channel identity. For every channel, the first 10~s snippet (from training dataset) is fed to the embedding encoder to produce its functional vector; this vector is then projected to the model width and reused unchanged for all tokens originating from that channel (across time patches and windows). We fuse the projected embedding with time–patch features and a time-only positional code to form the tokens consumed by the transformer. Variable channel counts are handled with padding masks.
\vspace{-2mm}
\paragraph{Masked-Region Reconstruction Setup.}
\vspace{-1mm}
For each window we have multichannel LFP signals $\mathbf X\!\in\!\mathbb R^{C\times T}$ and a per–channel
\emph{functional embedding} matrix $\mathbf E\!\in\!\mathbb R^{C\times F}$ with $F{=}32$.
At evaluation, a region $\mathcal R$ (e.g., VO) is fully masked: its channels provide no signal.
Let $\mathcal S=\{c:\,c\notin\mathcal R\}$ be \emph{sources} and $\mathcal T=\{k:\,k\in\mathcal R\}$ be \emph{targets}.
Given source signals $\{\mathbf x_c\}_{c\in\mathcal S}$ and source and target embeddings $\{\mathbf e_c\}_{c\in\mathcal S\cup\mathcal T}$, the model reconstructs target signals
$\{\widehat{\mathbf y}_k\}_{k\in\mathcal T}$.
\vspace{-2mm}
\paragraph{Functional tokens.}
\vspace{-1mm}
For each channel, a single 10\,s snippet (from training data) is fed to the embedding encoder to produce
its functional vector $\mathbf e_c\!\in\!\mathbb R^{32}$. This vector is linearly mapped to model width
$\mathbf f_c\!\in\!\mathbb R^{d}$ and then reused unchanged for every token originating from that channel
(across patches and windows). Layer normalization and learned type embeddings follow standard practice
and are omitted from equations for clarity.
\vspace{-2mm}
\paragraph{Tokenization of source signals.}
\vspace{-1mm}
A 1D conv ``tokenizer'' converts each source $\mathbf x_c$ into $P$ time patches. Let $t\in\{1,\dots,P\}$ index the patch.
Formally,
\[
\mathbf Z_c \;=\; \mathrm{Conv1d}(\mathbf x_c;\,L_{\text{patch}},S)\in\mathbb R^{P\times d},\qquad
P \approx \Big\lfloor \tfrac{T-L_{\text{patch}}}{S}\Big\rfloor + 1,
\]
where $d$ is both the number of conv filters and the Transformer width. $\mathbf Z_{c t}\in\mathbb R^{d}$ is the $d$-dim
temporal feature for channel $c$ at patch $t$.
We \emph{broadcast} the functional identity $\mathbf f_c\in\mathbb R^{d}$ to all patches of channel $c$ and
\emph{concatenate} it with the temporal features, $[\mathbf Z_{c t}\Vert \mathbf f_c]\in\mathbb R^{2d}$.
A dense layer \emph{reduces back to width $d$} and we add a \emph{time-only} positional code $\mathbf P^{(t)}_t\in\mathbb R^{d}$,
yielding a source token of shape:
\[
\widetilde{\mathbf Z}_{c t} \;\in\; \mathbb R^{d} \qquad (c\in\mathcal S,\; t=1,\dots,P).
\]
Stacking tokens across channels and patches produces the encoder input sequence
\[
\mathbf X_{\text{src}} \;=\; [\,\widetilde{\mathbf Z}_{c t}\,]_{c\in\mathcal S,\; t=1..P} \;\in\; \mathbb R^{S_{\text{tok}}\times d},
\quad
S_{\text{tok}} = |\mathcal S|\,P,
\]
with a key-padding mask $\mathbf m_{\text{src}}\in\{0,1\}^{S_{\text{tok}}}$ to ignore padded (absent) sources within a batch (due to variable number of channels present in the dataset).
\vspace{-2mm}
\paragraph{Tokenization of query signals (masked reconstruction targets).}
\vspace{-1mm}
Targets have no signal, so temporal features are replaced with a \emph{learned query base}
$\mathbf U_t\in\mathbb R^{d}$ (one $d$-vector per patch index $t$).
For each target channel $k\in\mathcal T$ and patch $t$, we concatenate the base with the functional embedding of that channel
$[\mathbf U_t \Vert \mathbf f_k]\in\mathbb R^{2d}$, reduce back to width $d$, and add the same time-only positional code,
producing query tokens of shape:
\[
\mathbf Q_{k t} \;\in\; \mathbb R^{d} \qquad (k\in\mathcal T,\; t=1,\dots,P).
\]
Stacking all targets and patches forms the decoder input sequence
\[
\mathbf Q \;=\; [\,\mathbf Q_{k t}\,]_{k\in\mathcal T,\; t=1..P} \;\in\; \mathbb R^{T_{\text{tok}}\times d},
\quad
T_{\text{tok}} = |\mathcal T|\,P,
\]
with a query-padding mask $\mathbf m_{\text{tgt}}\in\{0,1\}^{T_{\text{tok}}}$ to ignore padded (absent) targets.

\vspace{-2mm}
\paragraph{Encoder-decoder attention.}
\vspace{-1mm}
We use pre-LN Transformer blocks with $h$ heads and per–head size $d_h=d/h$. In particular multi-head self-attention (MHSA) and multi-head attention (MHA) blocks are used where :
\[
\mathrm{MHSA}(\mathbf X)=\mathrm{Concat}_h\!\big[\mathrm{Attn}(\mathbf Q_i,\mathbf K_i,\mathbf V_i)\big]\mathbf W^O,
\quad
\mathbf Q_i=\mathbf X\mathbf W_i^Q,\ \mathbf K_i=\mathbf X\mathbf W_i^K,\ \mathbf V_i=\mathbf X\mathbf W_i^V,
\]
where $\mathbf W^O$ projects $h$ concatenated outputs to model width ($d$), $i$ denotes a specific head from $h$ heads, and $\mathrm{Attn}$ defined as:
\[
\mathrm{Attn}(\mathbf Q_i,\mathbf K_i,\mathbf V_i)
=\mathrm{softmax}\!\left(\frac{\mathbf Q_i\mathbf K_i^\top}{\sqrt{d_h}}\right)\mathbf V_i.
\]
MHA follows identical formulation where query ($\mathbf Q_i$) comes from a source other than $\mathbf X$.

\textbf{Encoder (self-attn on sources).}\\
With source tokens $\mathbf X_{\text{src}}\in\mathbb R^{S_{\text{tok}}\times d}$,
\[
\underbrace{\mathbf Q^{\text{enc}}=\mathbf X_{\text{src}}\,\mathbf W^{Q}_{\text{enc}},\quad
\mathbf K^{\text{enc}}=\mathbf X_{\text{src}}\,\mathbf W^{K}_{\text{enc}},\quad
\mathbf V^{\text{enc}}=\mathbf X_{\text{src}}\,\mathbf W^{V}_{\text{enc}}}_{\in\ \mathbb R^{S_{\text{tok}}\times d}}
\ \Rightarrow\
\mathbf M=\mathrm{MHSA}(\mathbf X_{\text{src}})\in\mathbb R^{S_{\text{tok}}\times d},
\]
with key-padding mask $\mathbf m_{\text{src}}$ applied inside the softmax.

\textbf{Decoder (self-attn on queries, then cross-attn to memory).}\\
First, self-attention over query tokens $\mathbf Q\in\mathbb R^{T_{\text{tok}}\times d}$:
\[
\underbrace{\mathbf Q^{\text{dec}}=\mathbf Q\,\mathbf W^{Q}_{\text{dec}},\ \mathbf K^{\text{dec}}=\mathbf Q\,\mathbf W^{K}_{\text{dec}},\ \mathbf V^{\text{dec}}=\mathbf Q\,\mathbf W^{V}_{\text{dec}}}_{\in\ \mathbb R^{T_{\text{tok}}\times d}}
\ \Rightarrow\
\widetilde{\mathbf Q}=\mathrm{MHSA}(\mathbf Q)\in\mathbb R^{T_{\text{tok}}\times d},
\]
masking padded query tokens via $\mathbf m_{\text{tgt}}$. Next, \emph{cross-attention} where queries attend to encoder memory:
\[
\underbrace{\mathbf Q^{\text{qry}}=\widetilde{\mathbf Q}\,\mathbf W^{Q}_{\text{qry}}}_{\in\ \mathbb R^{T_{\text{tok}}\times d}},\quad
\underbrace{\mathbf K^{\text{mem}}=\mathbf M\,\mathbf W^{K}_{\text{mem}},\ \mathbf V^{\text{mem}}=\mathbf M\,\mathbf W^{V}_{\text{mem}}}_{\in\ \mathbb R^{S_{\text{tok}}\times d}}
\ \Rightarrow\
\mathbf H=\mathrm{MHA}(\widetilde{\mathbf Q},\mathbf M,\mathbf M)\in\mathbb R^{T_{\text{tok}}\times d},
\]
with key-padding mask $\mathbf m_{\text{src}}$ applied to the memory keys.

Stacking $L_{\text{enc}}$ and $L_{\text{dec}}$ layers yields the final decoder states $\mathbf H$.
\vspace{-2mm}
\paragraph{Waveform head and outputs.}
\vspace{-1mm}
A linear head maps each decoder token to a patch of $L_{\text{patch}}$ samples, then patches are reshaped and concatenated along time:
\[
\mathbf H \in \mathbb R^{T_{\text{tok}}\times d}
\;\xrightarrow{\ \text{linear}\ }\;
\mathbf R \in \mathbb R^{T_{\text{tok}}\times L_{\text{patch}}}
\;\xrightarrow{\ \text{reshape}\ }\;
\widehat{\mathbf Y} \in \mathbb R^{K\times P\times L_{\text{patch}}}
\;\xrightarrow{\ \text{concat over }t\ }\;
\widehat{\mathbf Y} \in \mathbb R^{K\times T_m}
\] 
\[,T_m=P\,L_{\text{patch}}\]
(An overlap--add variant with stride $S{<}L_{\text{patch}}$ is compatible but not used here.)
\vspace{-2mm}
\paragraph{Masking and batching.}
\vspace{-1mm}
For each window, sources exclude all channels in $\mathcal R$ and targets include all channels in $\mathcal R$.
Across windows/subjects the numbers of sources and targets vary; we pad to the batch maxima and use
$\mathbf m_{\text{src}}$ and $\mathbf m_{\text{tgt}}$ so attention and losses ignore padded tokens.
\vspace{-2mm}
\paragraph{Objective and optimization.}
\vspace{-1mm}
Given ground truth $\mathbf Y\in\mathbb R^{K\times T_m}$ and predictions $\widehat{\mathbf Y}$,
\[
\mathcal L
\;=\;
\underbrace{\tfrac{1}{T_m}\|\widehat{\mathbf Y}-\mathbf Y\|_2^2}_{\text{MSE}}
\;+\;
\lambda\Big(1 - \rho(\widehat{\mathbf Y},\mathbf Y)\Big),
\qquad \lambda=0.05,
\]
where $\rho$ is mean Pearson correlation across targets (after per-sequence standardization).
The correlation term enforces \emph{shape fidelity} (phase/temporal structure) and mitigates the tendency of pure MSE to prefer amplitude shrinkage or flat predictions under SNR or scale mismatch.
No subject IDs, subject-specific heads, or fine-tuning are used; a single model is trained and evaluated across all subjects for multi-subject experiments.

\vspace{-2mm}
\paragraph{Model parameters and training.}
\vspace{-1mm}
For training on multi-session multi-subject data, we use $L_{\text{enc}}=3$ encoder layers and $L_{\text{dec}}=3$ decoder layers with model width $d=128$ and $h=4$ attention heads ($d_h=32$). The Transformer feed-forward dimension is 384 with dropout set to $0.2$. The tokenizer patch length is $L_{\text{patch}}=25$\,ms. Each model input window is 1\,s of LFP, drawn from a variable number of channels (subject-dependent). Windows are extracted with a 500\,ms overlap after the train/validation/test split to avoid leakage. The resulting model has approximately $1.37\times 10^6$ trainable parameters.

Optimization uses AdamW with $\beta_1=0.9$, $\beta_2=0.99$, and $\varepsilon=10^{-8}$. We employ per-parameter learning rates: $2\!\times\!10^{-4}$ for all layers except the waveform head, which uses $3\!\times\!10^{-4}$. Weight decay is $10^{-4}$, except it is disabled for parameters that should not be regularized (e.g., LayerNorm scales/biases, embedding and type-embedding vectors, and the learned query base), following common practice. 150 epochs and batch size of 128 was used for training. 

For experiments trained on a single session (500–800 windows), we use a smaller Transformer with model width $d=32$ and feed-forward dimension $=32$, yielding $\sim 9.0\times 10^{4}$ trainable parameters. All other architectural and optimization settings remain the same, except that we increase the learning rates by $10\times$ (i.e., $2\times 10^{-3}$ for all layers and $3\times 10^{-3}$ for the waveform head). Batch size and number of epochs are identical to the multi-subject setting.

\subsection{Baselines and Associated Training Protocol}
\label{A_baseline}
We compare our transformer against a set of standard time–series baselines that map \emph{source} channels to \emph{target} channels within each window. For every subject and sessions, we use identical train/validation/test splits and the same masking pipeline: all channels whose region label matches a pre-specified exclusion predicate (e.g., \texttt{VO*}) are designated as targets, while the remaining channels are sources. 

\paragraph{Baselines.}
\begin{itemize}
\item \textbf{CopyBest} (nonparametric). For each target channel, we select the single source channel with the highest training correlation (Z-scored across time) and fit a scalar gain by least squares; the prediction is a scaled copy of that source. This yields a strong correlation-based lower bound without temporal modeling.

\item \textbf{FIR (linear, causal MIMO).} A multi-input multi-output finite impulse response filter that predicts targets from sources using a causal 1-D convolution with kernel size $L{+}1$. We set $L{=}64$ (thus receptive field $65$ samples equivalent to 65ms), implemented as a single causal Conv1d from $C_{\mathrm{src}}$ to $C_{\mathrm{tgt}}$ and trained end-to-end.

\item \textbf{TCN (temporal CNN).} \citep{BaiTCN2018} A temporal convolutional network with pointwise input/output projections and $D{=}5$ residual dilated blocks (dilations $1,2,4,8,16$), kernel size $k{=}5$, width $128$, GroupNorm(1) in each block, GELU activations, and dropout $0.3$. The effective receptive field approximately covers the FIR context. 

\item \textbf{GRU (recurrent).} A 2-layer Gated Recurrent Unit (GRU)~\citep{cho2014learning} with hidden size $128$ (dropout $0.3$ between layers), followed by a linear projection to $C_{\mathrm{tgt}}$ at each time step.
\end{itemize}

\paragraph{Optimization and early stopping.}
All trainable baselines (FIR, TCN, GRU) are trained with Adam \citep{KingmaBa2015Adam} (learning rate $3{\times}10^{-4}$, weight decay $10^{-4}$), batch size $8$, for up to $150$ epochs with early stopping (patience $10$) based on validation MSE.  

\paragraph{Evaluation metric.}
We report Fisher-$\bar r$ over all valid window$\times$target pairs:
\[
z_i=\operatorname{arctanh}(r_i),\quad
\bar z=\frac{1}{N}\sum_i z_i,\quad
\bar r=\tanh(\bar z),
\]
where $r_i$ is the per-sample Pearson correlation (over time) between prediction and ground truth for a target channel.

\paragraph{Statistics.}
For across-subject comparisons we used paired two-sided $t$-tests between each baseline and our method (FUNC2), applying both Holm--Bonferroni and Benjamini--Hochberg (FDR) corrections for multiple comparisons; all corrected $p$-values were $<\!0.001$, indicating statistically significant improvements.

\begin{figure}\center{\includegraphics[width=1\textwidth,clip=false] {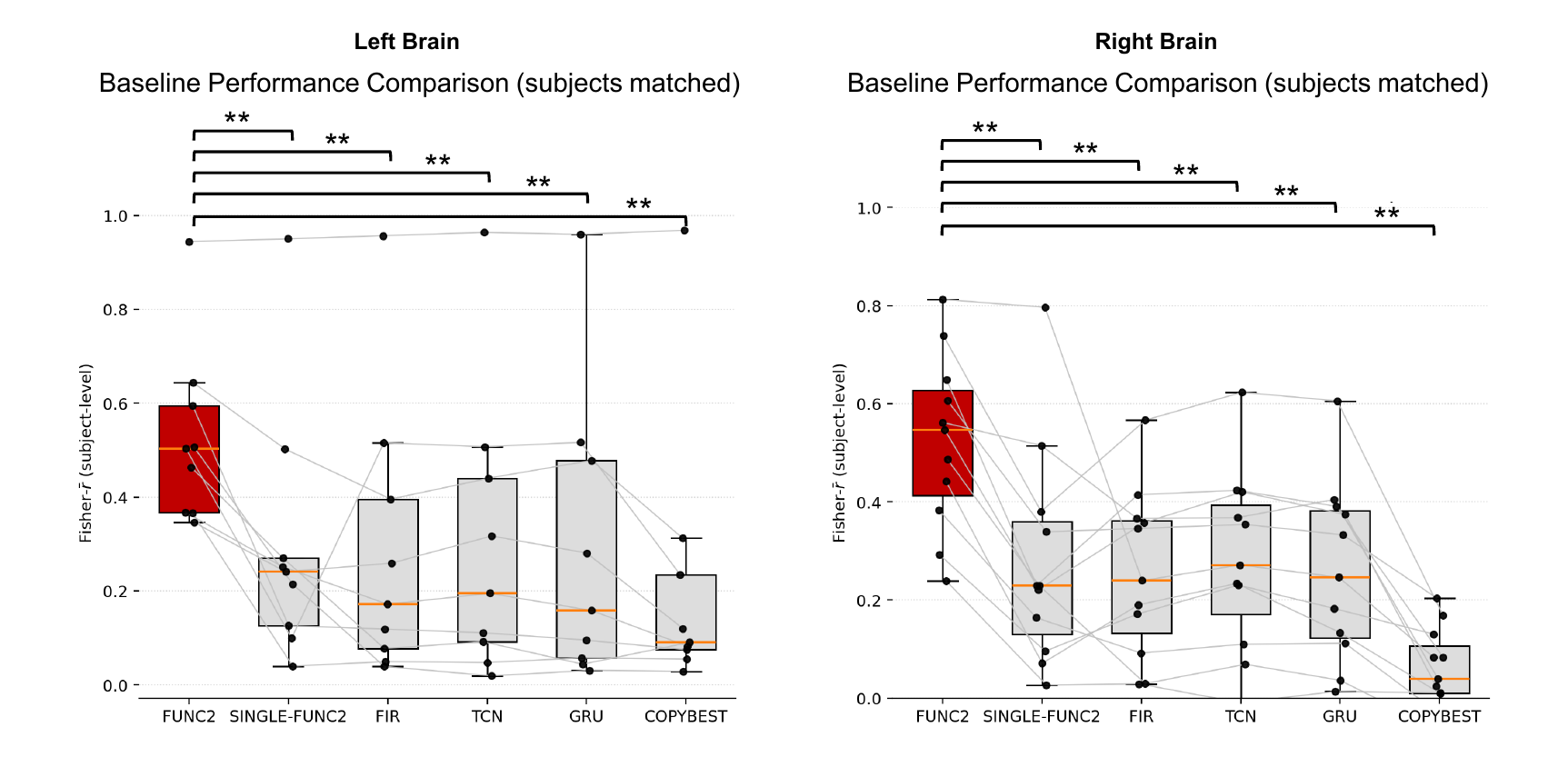}}
	\caption{\textit{Baseline comparisons (subject-level Fisher-$\bar r$).} Boxes show median and IQR; dots are individual subjects; gray lines connect the same subject across models. Models: FIR (linear causal MIMO), TCN, GRU, CopyBest, and our Transformer with Functional Embedding V2 (FUNC2). The pooled (multi-subject) FUNC2 model outperforms all baselines. Paired $t$-tests vs FUNC2 with Holm and BH/FDR corrections: all $p_{\text{corr}}<0.001$. (**)}

	\label{fig_baselines}
\end{figure}

\paragraph{Baseline comparison.}
Figure~\ref{fig_baselines} summarizes subject-level performance (Fisher-$\bar r$) for our proposed \textbf{Transformer+Functional~Embedding~V2 (FUNC2)} against standard time–series baselines: \textbf{FIR} (linear causal MIMO), \textbf{TCN}, \textbf{GRU}, and \textbf{CopyBest} (correlation-selected source with scalar gain). Each dot is a subject and gray lines connect the same subject across models (paired comparison). The pooled (multi-subject) \textbf{FUNC2} model clearly outperforms all baselines, indicating that the transformer with functional coordinates can effectively \emph{aggregate} cross-subject data to learn stronger source$\to$target mappings. We also include a \textbf{Single-FUNC2} control (the same transformer+functional embedding trained \emph{per subject}); as expected, it underperforms the pooled model because transformers are data-hungry and a single session in this dataset contributes only $\sim$500–900 windows—insufficient for reliable convergence. Statistical tests (paired two-sided $t$-tests vs FUNC2 with Holm and BH/FDR corrections) show all comparisons are significant ($p_{\text{corr}}<0.001$).

\subsection{Compute resources and training times}
\label{A_resources}

All models were trained on a single workstation equipped with an AMD Ryzen Threadripper PRO 7955WX (16 cores), 128~GiB RAM, and 1$\times$ NVIDIA RTX~4500 Ada (24~GiB VRAM). 
Table~\ref{A_table_resources} summarizes training times and data scale for each model. 
Notably, the \emph{Functional encoder (PSC, multi-subject)} is pair-based: given $n$ input samples, it can form $\mathcal{O}(n^2)$ pairs, enabling large-scale training (here, $2{,}500{,}000$ pairs). 
In contrast, \emph{Functional encoder (MSC, multi-subject)} computes its loss over full sample batches (not pairwise), so the effective sample count remains at $65{,}000$ despite the same source dataset.\\

\begin{table}[h]
    \caption{\textit{Training times and data scale.} Reported on a single workstation (AMD Ryzen Threadripper PRO 7955WX, 128~GiB RAM, 1$\times$ NVIDIA RTX~4500 Ada 24~GiB).\\}
    \label{A_table_resources}
    \centering
    \small
    \begin{tabular}{lccc}
        \textbf{Model} & \textbf{Epochs} & \textbf{Samples} & \textbf{Time} \\
         \hline
        Functional encoder (single-subject) 
            & 100 
            & 20{,}000 
            & $\sim$7 minutes \\
        Functional encoder (PSC, multi-subject) 
            & 30 
            & 2{,}500{,}000 
            & $\sim$5 hours \\
        Functional encoder (MSC, multi-subject) 
            & 200 
            & 65{,}000 
            & $\sim$1.5 hours \\
        Masked reconstruction transformer 
            & 150 
            & 27{,}000 
            & $\sim$2.5 hours \\
    \end{tabular}
    
\end{table}

\subsection{Large Language Models (LLMs) usage}
We used ChatGPT (OpenAI GPT-5) as a general-purpose assistant during the preparation of this work. The model was employed to help with editing for clarity, conciseness, and formatting. All scientific ideas, analyses, experiments, and interpretations presented in this paper were conceived, implemented, and validated by the authors. The authors take full responsibility for the content of this manuscript.

\end{document}